# Ben-Gurion University of the Negev

Faculty of Engineering Sciences
Department of Industrial Engineering and Management

## Computer vision system to count crustacean larvae

By: Chen Rothschild

Academic guidance: Prof. Yael Edan, Prof. Amir Sagi

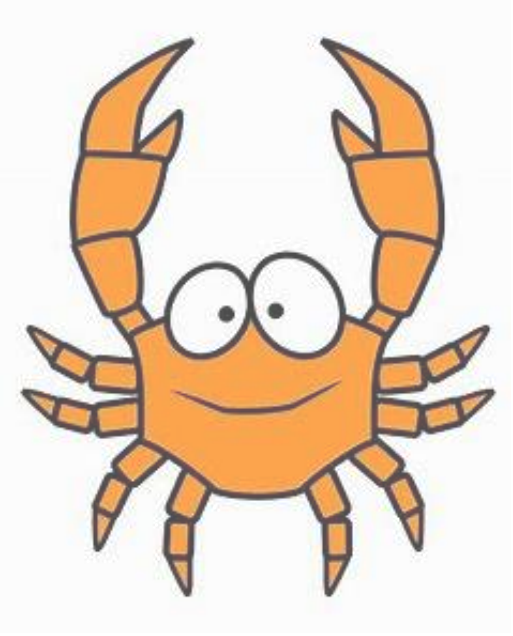

September 2022

# Ben-Gurion University of the Negev

Faculty of Engineering Sciences
Department of Industrial Engineering and Management

## Computer vision system to count crustacean larvae in industrialized ponds

By: Chen Rothschild

Academic guidance: Prof. Yael Edan, Prof. Amir Sagi

Author: Date:

Supervisor: Date:

Supervisor: Date:

Chairman of Graduate Studies Committee: Date:

September 2022

# Abstract

Fish products account for about 16 percent of the human diet worldwide, as of 2017. The average annual increase in global food fish consumption from 1961 to 2017 was 3.1 percent, nearly twice the growth rate of the world population (1.6 percent) and more than the growth rate of all other animal protein foods (e.g., meat, dairy, milk).

The counting action is a significant component in growing and producing these products. Growers must count the fish accurately mainly to manage an accurate feeding strategy and plan a schedule and quantities for marketing and sales. to do so technological solutions are needed. Digital aquaculture relies heavily on intelligence technologies like computer vision to increase product quality and efficiency.

Two computer vision systems to automatically count crustacean larvae grown in industrial ponds were developed. The first system included an iPhone 11 camera with 3024X4032 resolution which acquired images from an industrial pond in indoor conditions with two different illumination conditions; a 16-liter white bucket was specially placed inside to provide a good background. Two experiments were performed with this system- the first one included 200 images acquired in one day on growth stages 9-10 with an iPhone 11 camera on specific illumination conditions. In the second experiment, a larvae industrial pond was photographed for 11 days with two devices- an iPhone 11 camera and a SONY DSC-HX90V camera. With the first device (iPhone 11) two illumination conditions were tested. In each condition, 110 images with a resolution of 3024X4032 were acquired.

The second system included a DSLR Nikon D510 camera with a 2000X2000 resolution with which seven experiments were performed outside the industrial pond. Images were acquired on day 1 of the larvae's growing stage resulting in the acquisition of a total of 700 images.

An algorithm that automatically counts the number of edible crustaceans was developed for both cases based on the 'YOLOv5' CNN model. A major challenge in this study was the small size of the objects.

The system for counting larvae inside the industrial ponds resulted in an accuracy of 88.4% image detection and mAP of 0.855 with an iPhone 11 camera and illumination condition number two including all the images from all days.

The second system for counting larvae outside the industrial ponds resulted in an accuracy of 86% and mAP of 0.801 for a density of 50. When applying the combination density method, average identification accuracies improved in all densities, and to 88% for density 50.

In aquaculture, fish growth models are crucial to predict fisheries outcomes, improve fish farms, and analyze the dynamics of fish populations. In this study, a growth function for Macrobrachium Rosenberg's larvae was developed. The experiment included 10 larvae for each growth stage, out of 11 growth stages for 19 days. Every stage is characterized by physical changes in larvae that are mainly expressed in the eyes, rostrum, antennal flagellum, uropod, telson, pleopods, and pereopods. The experiment was located at Ben-Gurion University's life science lab.  An image or several images of every larva were taken manually from the industrial pond. Daily, several larvae were taken manually from the industrial pond and analyzed under a microscope -Nikon ECLIPSE Ci-S equipped with a Nikon DS-Fi3 RGB camera in the resolution of 2880X2048. The aim was to determine their growth stage. Once the growth stage was determined, an image or several images of the larva were acquired. Each larva's length was measured manually from the images.
The final step was fitting a growth function to the data. After fitting 5 different models, the most suitable model was chosen to be the Gompertz model with a goodness of fit index of R^2=0.983.
**Keywords:** Computer vision, Larvae, YOLOv5, algorithm, object detection, growth model.

Table of Contents

# 1 Introduction

## 1.1 Problem description

Fish contribute a significant amount of protein to a person's diet. Over the years, the demand for world capture fisheries and aquaculture production increased dramatically - from 20 million tons of fish consumed in 1950, to about 180 million tons in 2018 (FAO, 2020). As of 2017, fish products constitute about 16% of the human diet around the world (Ibrahin et al., 2017). Hence, it is necessary to increase the number of fish for food consumption purposes and to ensure fish species multiply with maximum efficiency (Costa et al., 2019).The aquaculture industry can provide increased fish production along with improved quality (Ibrahin et al., 2017).
The counting operation is an important process in automatic fish production (Ibrahin et al., 2017) and a main problem involved in raising small aquatic organisms (Chatain et al., 1996).
The number of fish is an important parameter for fish production to determine exactly the size and quantity of cages/ponds for production; to plan marketing schedules and to manage accurate feeding strategies. In most fisheries, fish counting is done manually, a time consuming process requiring intensive human labor ( Albuquerque et al., 2019). In addition, manual counting is subject to error and omission due to fatigue and inattention ( França Albuquerque et al., 2019). Furthermore, manual counting can lead to fish stress resulting sometimes even in fish death.
Aquaculture production has been automated using computer vision (Le & Xu, 2017). Digital aquaculture benefits greatly from the application of intelligence technologies to improve product quality and production efficiency (Yang et al., 2021). Fast, accurate, and efficient fish counting is essential (França Albuquerque et al., 2019), and counting based on image processing can meet these need (Le & Xu, 2017). Computer vision in agricultural has been applied to a wide range of applications (provide references) due to its many benefits, such as good accuracy, low expense, and high reliability (Costa et al., 2019).
A convolution neural network (CNN) is a popular deep learning technique (Yang et al., 2021). With CNN, fish species can be identified, detected, and located with high accuracy (Yang et al., 2021). When using neural networks it is important that the network will be train on a sample of the population and generate the correct response to each input (Newbury et al., 1995). In machine learning and computer vision, small objects pose a challenge because they take up less space in the image and contain less information for models to learn from (Babu et al., 2022) (Bochkovskiy et al., 2020). Even so, it is possible for computer vision to count large and small objects (Fan & Liu, 2013).

Fish growth reflects the precision and efficiency of aquaculture production (Lugert et al., 2016). An increase in a living system over time is referred to as growth, growing fish in aquaculture production systems differs from growing wild fish, which is why it is crucial to know the limits of growth before breeding (Baer et al., 2011). Macrobrachium rosenbergii growth is primarily affected by temperature, salinity, pH, light intensity, and feeding conditions (Zhang et al., 2006).

The growth performance of organisms in commercial aquaculture facilities is the most important influencing factor regarding economic benefit (Baer et al., 2011).

## 1.2 Objectives

The research objective is to develop a computer vision system that will automatically count the number of larvae in a growing industrial pond and develop a larvae growth function. Specific objectives are to:

1. Develop and evaluate computer vision system for counting larvae in indoor controlled ponds along the growth period:
    a. Design the imaging system – camera location, illumination
    b. Develop algorithm
    c. Evaluate performance for different growing days and conditions
2. Develop and evaluate algorithms for counting larvae from growth stage 1 under microscope conditions
3. Develop growth function for larvae and develop/evaluate ability of automatic machine vision system

# 2 Literature Review

## 2.1 Background

Agriculture and its sub is a major source of nutrition in daily life for about 60% of the world's population (Mim et al., 2020). Furthermore, agriculture is a major source of livelihood and it is used to increase the income of countries around the world (Gobalakrishnan et al., 2020). Farm crops include vegetables, fruits, fish, and animals, grown to meet these need (Mim et al., 2020). The world's population is expanding, leading to a reduction in cultivated land and an increase in urbanization (Tian et al., 2020), along with an increasing need for food (Mim et al., 2020). Hence, the need for plots will continue to increase (Tian et al., 2020), as well as the demand for effectual food production methods (Tian et al., 2020).

Fish are an essential source of protein, and as a result, has become a major component of human nutrition all over the world (Li et al., 2020). Aquaculture is the fastest-growing food sector in the world (FAO, 2020). According to the Food and Agriculture Organization of the United Nations (FAO), in 1961 global fish consumption was 9 kg/person per year while in 2016 it increased to 20.3 kg/person per year (Costa et al., 2019). Precise aquaculture is a real alternative to the harvesting and consumption of wild fish (Antonucci & Costa, 2020). Due to this fact, it is necessary to increase the number of fish for food consumption purposes and to ensure that fish species multiply with maximum efficiency (Costa et al., 2019). To do so and to increase production and ensure fish quality, while maintaining its well-being, it becomes even more important to monitor and control the production process (Antonucci & Costa, 2020). With the development of information and intelligent systems, aquaculture is increasingly prone to industrial culture – (Li et al., 2020) reducing inputs (such as feed), optimizing outputs, and decreasing pollution (Antonucci & Costa, 2020). With the development of new technologies, researchers and practitioners of aquaculture have explored various methods without hand-operated intervention (Li et al., 2020). The basis of precise aquaculture is the application of better control, monitoring, and documentation principles of biological processes in fish (Antonucci & Costa, 2020). Technological developments like robotics, computer vision and, machine learning have proven to improve agriculture production (Redolfi et al., 2020).

## 2.2 Image processing in agriculture

Computer vision uses a camera and a computer instead of the human eye to make a machine "see" (Tian et al., 2020). A computer vision system uses a camera which can be RGB (red-green-blue) for visual inspection, multi-spectral, or hyper-spectral range sensors for geometric estimation (Lu & Young, 2020). Image processing techniques and machine learning technologies are benefiting farmers in varied aspects of agricultural automation production management (Manohar & Gowda, 2020), such as growth monitoring, disease prevention, and fruit harvesting (Tian et al., 2020). By reducing manual labor and production costs, increasing yield and quality, and improving control, agricultural automation has been able to increase productivity many folds Concerned about environmental implications (Kapach et al., 2012).Computer vision helps improve speed and accuracy (Reddy et al., 2020) in a variety of tasks in the life cycle of the plant - from planting to harvest (Lu & Young, 2020). By using image processing methods, we can turn the information into knowledge and use it (Figure 1) leading to improved decisions (Tombe, 2020).

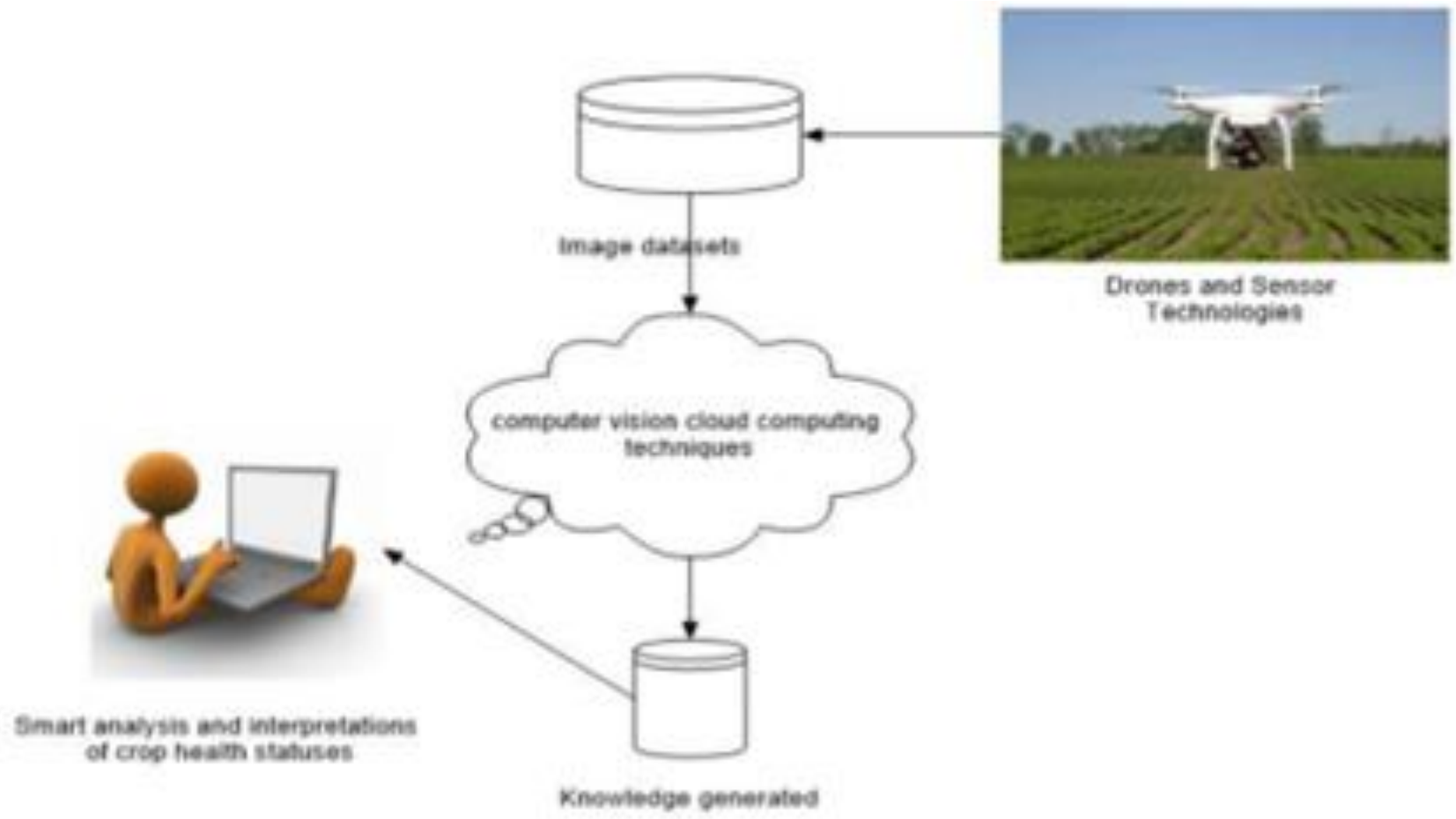

Figure 1- The management cycle of smart farming (Tombe, 2020)

For example, some algorithms accurately identify various infections in plants (Gobalakrishnan et al., 2020), and some classify the type of soil according to sampling (Deorankar & Rohankar, 2020). Along with the many benefits of computer vision in agricultural, such as good accuracy, a wide range of uses, low expense, high reliability, and unlike the human eye it never gets tired (Costa et al., 2019), there are also challenges:

- Professionalism - The more widely used image processing technology, the higher the requirements for professionals and professionals in the field. To develop new technologies, explore and integrate solutions from various fields of agriculture (Tian et al., 2020).
- Choosing an algorithm - In the application of computer vision technology, a variety of characteristics must be considered to select the appropriate algorithm. There is no default application of computer vision technology. In any case, the properties depend on the environment and target. which in agriculture are characterized by high variability and are not structured. (Tian et al., 2020).

- Work environment - Most existing technologies and methods are implemented in a laboratory environment or experimental platform. But the data in the natural environment differs from the data in the experimental environment, due to natural factors (Tian et al., 2020) with changes along time and space.
- Image size - When air images are used for example, some images include sizes exceeding 20000 × 30000 pixels. Such sizes are difficult to process in terms of calculation time and memory space (Chiu et al., 2020).

## 2.3 Counting in agriculture base on image processing

Improving crop productivity become a critical task these days, as the world population has grown exponentially (Fuentes-Penailillo et al., 2019). As the population grows, so does the global demand for food security. In addition to food consumption, agricultural produce plays an important role in our daily lives - it is used for chemical or industrial products, cosmetics, furniture, etc. (Errami & Khaldoun, 2018). To achieve this goal, the various activities carried out in the growing area must be monitored efficiently (Fuentes-Penailillo et al., 2019).

While the current methodology involves a manual process and usually takes more than a week, an automatic counting technique can be developed using a camera system and image processing technology (Padilla et al., 2019). Manual counting is tedious work, requiring extensive human resources and cost, and providing low accuracy. Accurate counting helps farmers plan the manpower for harvesting, shipping, sales, and harvest-related operations. Computer vision techniques can help count accurately (Ri & Xvlqj, 2019). Accurate and reliable counting of fruit and vegetable counts from photos, for example in orchards, is a challenging problem that has received significant attention recently. Pre-harvest fruit count assessment provides useful information for logistical planning, enabling farmers to track their current orchard status, plan for optimal utilization of their labor force, and make informed decisions before the harvest process (Häni et al., 2018). In addition the counting is also important for studies dealing with the collection of yield-related phenotype data (Häni et al., 2018). These numbers are now estimated by manually sampling several trees at random and making an assessment based on this, for the entire orchard. This count, of course, is inaccurate (Häni et al., 2018). Precise farming techniques have received much attention as they provide a solution that can maximize yield while reducing economic and environmental costs in production. By applying the exact amount, the yield can be maximized while reducing the application of fertilizers and pesticides (Errami & Khaldoun, 2018). Therefore, fruit counts has been developed using in-field machine vision (Wang et al., 2017). The following table (Table 1) reviews several studies that present methods for counting using image processing and various algorithms.

Table 1- Counting applications in agriculture

| Objective | Object | Sensors | Algorithm | Experimental conditions | Data | Results | Reference |
|---|---|---|---|---|---|---|---|
| Develop methods to automatically, and accurately detect and count citrus fruits. | Citrus fruits | Thermal camera FLIR® A35s with a micro bolometer detector VOX type uncooled. sensitivity 0.05 °C, fixed focus, resolution of 320 x 256 pixels a spectral range from 7.5 to 13 nm, an accuracy ±5 °C or ±5% in reading and a dynamic range of 14 bits. | MATrix LABoratory, segmentation using Gamma function and fruit detection using Circle Hough Transform. | -Location: field 111 of Maringa´ Farm, located in Gaviao Peixoto – SP.<br>- The images were taken at different times of day (morning, afternoon, and evening). | Ground troth:<br>98 trees, the trees were the variety - Ruby and they were planted in November 2010. | Accuracy:<br>1. isolated fruits- 93.5%<br>2. Occluded fruits- 80%. | (Pedraza et al., 2019). |
| Develop a method that can accurately estimate the soybean seed number from an individual pod image | Soybean seeds- | iPhone 8 camera with 12 million pixels, f / 1.8 aperture, manufactured by Apple Inc. set the focal length to 4 | TCNN -Two column Convolution Neural Network. | The database images were taken during daytime under natural light conditions during mid to late October at | A total of 32126 seeds with their centers annotated, - different varieties of seeds. | MAE- 13.21<br>MSE- 17.62 | (Li et al., 2019). |
| with a single perspective. | | mm, ISO speed to 40, turn off the flash and live, turn on the anti-shake function. | | the Beijing Shunyi Base of the Institute of Crop Sciences, Chinese Academy of Agricultural Sciences. | | | |
| An automated olive tree counting method based on image processing of satellite imagery. | Olive trees | SIGPAC viewer of Ministry of Environment and Rural and Marine Affairs Spain. Resolution of 300 × 300 pixels. | 1.pre-processing using greyscale converted images and enhancing them using unsharp masking.<br>2. image segmentation through multi-level thresholding.<br>3. detection using circular Hough. Transform<br>4. counting using the blob analysis technique. | Imagery from over 16 Spanish communities. | 95 sample images with varying distribution of ground classes. | Overall accuracy of 96%. | (Khan et al., 2018). |
| A real time system that successfully detect trees in a scene and | Palm trees | Unmanned airborne vehicle (UAV) equipped with Sony cyber- | Hough transform. Blob detection-using the function | The study area is a periodic field of palm trees located in marrakesh, | 46 palm trees per 900 m2. | The accuracy of the proposed method | (Errami & Khaldoun, 2018). |
| also recognizes them in each and every frame. | | shot DSC-H90 16.1 MP RGB digital camera. | cvRenderBlobs. Feature extraction-algorithm known as the Center Surround Distribution Distance (CSDD). | morocco. | | obtained is 93%. | |
| Perform population counts in sunflower plants. | Sunflowr plants | Unmanned aerial vehicle (UAV) model DJI Phantom 3 Advanced equipped with a 12 megapixels camera was used to obtain images with a resolution of 4000×3000 pixels. The camera was set to speed priority with auto adjustment of up to maximum ISO = 1600. | 1. Image processing- using the Agisoft Metashape software.<br>2. Data processing- using the open source software R.<br>3. Extraction of subsectors, classification and counting of objects- k-means To each of the plots. The cluster corresponding to crop was processed using R package EBImage. | The experiment was carried out over a sunflower crop located in San Clemente, Maule, Chile during the 2018/2019 growing season. During the experiment, the crop was in its first stages of post-emergency development. | In the experimental site, a total of 4 plots were made. | Average error:<br>Vγ- 10%<br>VVARI-9%<br>standard deviation:<br>Vγ- 3%.<br>VVARI-9%. | (Fuentes-Penailillo et al., 2019). |
| Formulate fruit counting from images as a multi-class classification problem. | Apple | Standard Samsung Galaxy S4 cell phone camera. | Deep Neural Networks, CNN- AlexNet. | The images collected from multiple orchards in Minnesota and Washington. | Many different kinds of apple tree species. <br>Training set: <br>two datasets, one containing green and one containing red apples. Both datasets were obtained from the sunny side of the tree row. 13000 image patches and 4500 patches at random that do not contain apples. <br>Test set: <br>Total 4 test sets of 5054, 599, 574 anf 704 images of red, yellow and orange apples. Contains shadow as well as in bright illumination. <br>validation set: <br>1- six trees in a row with a total of 270 apples. <br>2- eight trees in a row with a total of 274 apples. | accuracies between 80% and 94%. | (Häni et al., 2018). |
| A novel approach, for detecting and counting, ripened chili fruit from the | Chili crop | Sony 7.1 Megapixel camera (model Number: HDR-PJ50E, Sony, China). | Delta E and Lab color. | The images captured on seven different days in a field grove in the Ginjupalli, | Thirty-one images of fruit variety named Byadagi. | 92.7097% accuracy. | (Ri & Xvlqj, 2019). |
| plant images using Image processing techniques. | | | | Atchampet mandal, Guntur district, Andhraprades, India in natural daylight condition during the month of February 2019. | | | |
| To apply machine learning (ML) techniques, more specifically, deep learning (DL) to measure the final population of plants in corn crops. | Corn plant | The UAV was the DJI Phantom 4 Pro. Its RGB camera has a 1-in CMOS sensor and 20 megapixels. | CNN, architecture and blob detection methodologies. | The fields located in the Triângulo Mineiro region of Minas Gerais in Brazil and were composed of corn crops. | 50 images, of which 25 were used for training, four for validation, and 21 for testing purposes. The training data set is composed of 513 labels created for corn plants and 215 for ground, while for validation, a total of 121 labels for corn plants and 56 labels for ground were created. | The training was finalized with a training loss of 0.03 and an accuracy of 0.994. | (Kitano et al., 2019). |
| Detect and estimate the weight of an individual melon automatically. | Melons | A “Phantom 4 Pro” UAV equipped with a color RGB DJI FC6310 type camera. The UAV hovered about 15 m above the field, with | RetinaNet network. | The data were acquired at midday on 23 July 2018 from a 180 × 260 m open field at Newe Ya’ar (32°43’05.4”N 35°10’47.7”E) in Israel. | 116- from the first dataset<br>32- from the second dataset. | overall average precision score of 0.92. | (Kalantar et al., 2020) |
| | | the camera facing vertically downward during the image acquisition. Image resolution was 5472 × 3648 pixels, with a ground sampling distance (GSD) of 4 mm width rectangular. additional data collected the year before from the same place with a Sony ILCE-5000 camera mounted on a Quad-Copter drone were also used. Image resolution was 5456 × 3064 pixels with a GSD of 5 mm × 4.2 mm. This data included 32 measured | | | | | |
| | | melons. | | | | | |
| A system which locates, tracks and estimates the ripeness and quantity of fruit in the field. | Sweet peppers | Intel RealSense SR300 | Parallel-RCNN | Imagery of sweet peppers were cap-tured from a commercial protective cropping structure located in Giru, North Queensland. The data was acquired over three days. | Train set- 152 images, 958 peppers in total. Eval set – 133 images, 923 peppers in total | F1-Score of 77.3. | (Halstead et al., 2018) |
| Leaf counting. | Leaf | **- Unspecified** | Direct regression and counting by detection of key-points : CNN- using FPN and ResNet-50 as a backbone. | **- Unspecified** | The LCC data set. A1 128 images, A2 31 images, A4- 624 images, all of them are images of Arbidopsis plant. A3- 27 images of young Taboco plants. The images size was 800×W where W is chosen to keep the original image aspect ratio. | 95% average precision. | (Farjon et al., 2020) |

## 2.4 Implementation of image processing in aquaculture

Image analysis has been extensively used in aquaculture (Antonucci & Costa, 2020). Visual inspection of animals can provide a variety of knowledge regarding their health and development (Antonucci & Costa, 2020), For example, their number, size, gender, and quality. Their well-being and behavior can also be monitored (Zion, 2012). In general, computer vision technologies allow for constant monitoring of animal welfare, conditions, and their environment as part of aquaculture while reducing waste use (Antonucci & Costa, 2020). Generally, a computer vision in aquaculture system starts with collecting, continues with processing and analyzing, and ends with reasoning, as shown in Figure 2 שגיאה! מקור ההפניה לא נמצא.

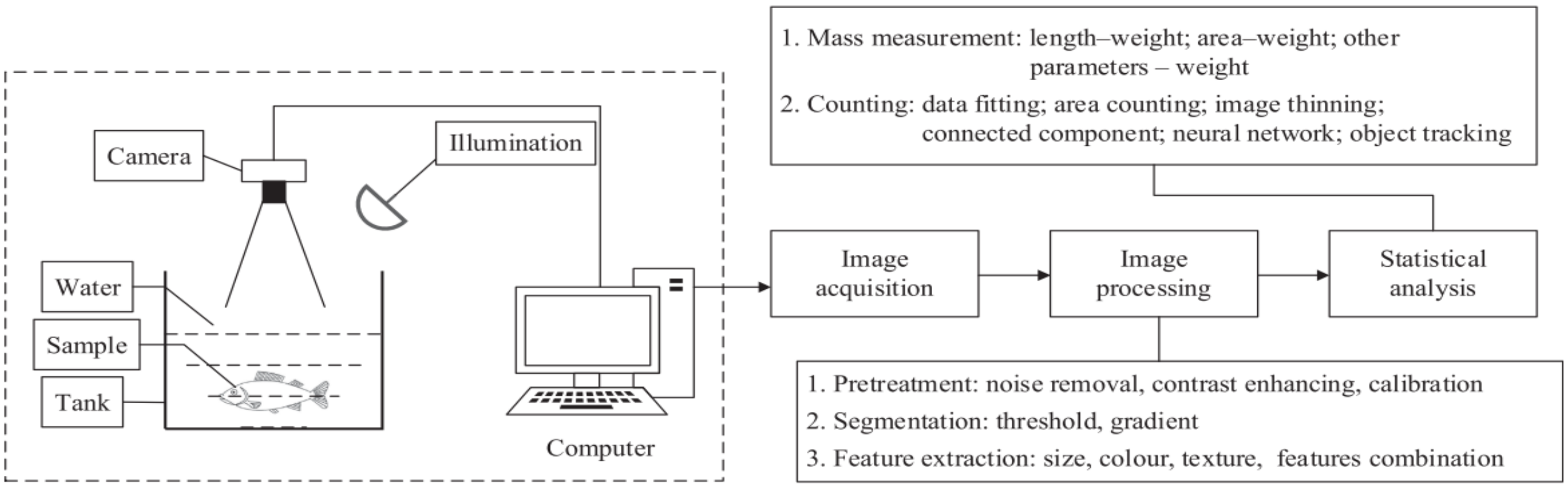

The application of computer vision technologies in aquaculture involves several complications - the tested subjects (fish) are sensitive to touch and movement. They are prone to stress and free to move in their natural environment, an environment where lighting cannot be controlled, and stability in most cases. Finally, the equipment must operate in a wet environment and is therefore likely to be expensive (Zion, 2012). Thus, these technologies must overcome limited visibility, lighting variations, changing distances and directions between cameras and objects, and even physical instability (Antonucci & Costa, 2020).

## 2.5 Livestock counting

Table 2- Livestock counting based on image processing

| Objective | Object | Sensors | Algorithm | Experimental conditions | Data | Results | Reference |
|---|---|---|---|---|---|---|---|
| Pig counting | Pigs | **- Unspecified** | Counting CNN and ResNeXt model. | Images from internet websites | 3401 images | MAE of 1.67 | (M. Tian et al., 2019) |
| Estimate the number of hens in battery cages | Hens | Off-the-shelf Media-Tech W9R (Action Camera W9R; SZ Ravo Technology Ltd, Shenzhen, China) with a 170-degree wide-angle lens | Faster R-CNN and NMS | The system was tested in 1 hen house of battery cages. An average of 71 frames per cage was obtained, at a camera speed of 30 fps. Data were collected on two different days | 5048 images, on average, there were 24.1 hens per cage | 9.6%accuracy and a mean absolute error (MAE) of 2.5 hens per cage | (Geffen et al., 2020) |
| Estimate the number of cattle and sheep | Cattle and sheep | The drone was equipped with an integrated Pan-Tilt- Zoom (PTZ) camera. The camera has a 2.3−1 -inch CMOS image sensor that can rotate flexibly in the | Mask R-CNN for feature extraction and training in the images captured by quadcopters | The dataset was collected from a private farmland in Armidale in Australia. The observation videos of 10 flight campaigns for livestock were | The dataset consists of 1000 images with 3737 livestock in total | Accuracy of 94.7 in cattle, 97.3 in sheep-96.0 in total. | (Xu, Wang, Falzon, Kwan, Guo, Sun, et al., 2020) |
| | | lateral and vertical | | recorded by the MAVIC PRO drone from April to October | | | |
| The feasibility of tracking and counting cattle at the continental scale from satellite imagery | Cattle | WorldView3 (Maxar) imagery of Amazon ranches, with panchromatic (PAN) and multi-spectral (MUL) bands at 0.31 and 1.24 m/pixel respectively | Compare between ImageNet pretrained-CSRNet and -LCFCN and an LCFCN with weights initialized using the Xavier | The top selling ranches in Brazil | 903 images containing a total of 28498 cattle | F1-Score of 0.676 in LCFCN, 0.571 in CSRNet | (Laradji et al., 2020) |
| Cattle counting in different situations | Cattle | MAVIC PRO drone which is equipped with an integrated PTZ camera shown in Fig. 3. The camera has a 1/2.3-inch CMOS image sensor that can rotate flexibly both laterally and vertically | Mask R-CNN | Used a drone to collect representative image data sets both in extensive pasture and in feedlot environments separately. The datasets were collected from the Tullimba Research Feedlot owned by the University of New England, | Each of the datasets both in the pastures and feedlot contain a total of 750 images | Accuracy of 94% in counting cattle on pastures and 92% in feedlots | (Xu, Wang, Falzon, Kwan, Guo, Chen, et al., 2020) |
| | | | | New South Wales, Australia, and surrounding farmlands across seasons from Summer to Spring (February to October) | | | |

## 2.6 Fish counting

In recent years, with the rise in popularity of consumers' perception of aquaculture crops, the aquaculture industry has developed rapidly (Le & Xu, 2017). Fish counting is an important and necessary process in the breeding process and should be done quickly, reliably, and efficiently. Most fishermen use manual fish counting in various forms, which requires intensive human labor which is time-consuming (França Albuquerque et al., 2019). Besides, the count is also used in the sale process itself - in fish farms and shops, nutritious fish, for example, are usually packaged for sale in plastic bags. During the packing process, the fish must be counted to know the exact quantity sold to customers. And as mentioned, as already mentioned this process is time-consuming and may include human errors (Toh et al., 2009). Compared to manual counting, beyond the fact that automatic counting reduces costs and time, it also improves counting accuracy (França Albuquerque et al., 2019). As technology improves, more and more industries are using computer vision and image processing technology to improve the level of automation of production. Respectively, fish counting based on image processing attracts more attention due to its high efficiency and accuracy (Le & Xu, 2017). **שגיאה! מקור ההפניה לא נמצא.** below describes fish counting methods based on image processing.

Table 3- Fish counting based on image processing

| Objective | Object | Sensors | Algorithm | Experimental conditions | Data | Results | Reference |
|---|---|---|---|---|---|---|---|
| The fish population is accurately counted using the number of endpoints. | Free-swimming fish | Camera- SHARP 1/4 CCD model, surrounded by a circle of LED compensating underwater light in the dark condition. The resolution of all the video- *640 x 480*. | Improved Otsu algorithm - skeleton extraction method. | Transparent glass aquarium- *1.2m x 0.6m x 0.5m.* | Free - swimming fish. hundreds of underwater images from the videos.<br><br>The exact number of images is **Unspecified**. | Average counting error less than 6%. | (Le & Xu, 2017). |
| Develop a simple photographic method for counting, of larvae and to compare its precision and efficiency with that of hand counting and | Sea bass fry | Polaroid 670 AF camera. | Sampling strategy: number of squares - simple stratified random sampling counting- optimal allocation | White plastic *85 cm X 85 cm X 15* cm sampling tray. The water depth should not exceed 40-50 mm. | Mean load of 2000- 4000 fish per tray. 50-day-old sea bass fry. Total-100000-150000 fish. | Error rate of 1.7%. | (Chatain et al., 1996). |
| weighing methods for fry of a few grams weight. | | | principle, Cochran (1963). | | | | |
| counts fish by combining information from blob detection, mixture of Gaussians and a Kalman filter. | Fingerlings - Pintado real | Logitech C920 PRO WEBCAM HD camera, with a resolution of 640x480 and a frame rate of 30 frames per second. | Background subtraction technique of Gaussian mixtures (MoG), techniques of segmentation, contour detection techniques, blob detection, and Kalman filter. | Two LED light bulbs positioned alongside each other. | 20 videos, where divided to 3 groups. | Average precision- 97.47%. | (França Albuquerque et al., 2019). |
| An automated system for counting fish by species. | Northern pike and the Atlantic salmon | Infrared fish silhouette sensor "RiverWatch".<br><br>Exact details are **Unspecified**. | The algorithm works in three steps: feature extraction, classification, and classifier fusion. Classification- combined results of a Bayes maximum likelihood classifier, a Learning Vector | Outdoor- the fishway at the HydroQuébec's Rivière-des-Prairies dam near Montreal. | Five specific species of fish.<br><br>Exact details are **Unspecified**. | Recognition rate of around 80%. | (Cadieux et al., 2000). |
| | | | Quantification classifier, and a One-Class-One-Network (OCON) neural network classifier. | | | | |
| A simple method of counting feeder fish automatically using image processing techniques. | Feeder fish | Video camera. **Unspecified**. | The median area. | Plastic bag. | One species of fish which look similar in shape and size to each other. | Percentage Error: 15 fish- 2.80% 50 fish- 3.38%. | (Toh et al., 2009). |
| A new automated fish counting system by using the principle of PTV analysis technique and image processing. | Japanese rice fish (Oryzias latipes) | Video camera- PIXPRO SP360, which has a 360° spherical curved lens and can record a 360° high definition (HD) image without the need for multiple cameras. | 1.image processing - background subtraction and the RECC<br><br>2. PTV analysis - KC Test. | Water tank- direct- *640 mm*, height - *460 mm.* volume -*100 L*. The tank was covered with glass wool to exclude surrounding noise. | 250 Japanese rice fish (Oryzias latipes) whose average total length is 2.99 cm. | Average Error- 8%. | (Abe et al., 2017). |
| This paper presents an accurate and automatic algorithm to recognize and count fish in the video footages of fishery operations. | Tuna, swordfish and mahi-mahi. | Video camera-resolution 640×480 | 1.machine learning algorithms-HLS, artificial neural network (ANN)<br>2. A statistical shape model (SSM) | An outdoor shipborne video system in an uncontrolled illumination environment. | 7 fish from 3 species- tuna, swordfish and mahimahi | Recognition accuracy is 89.6% | (Luo et al., 2016). |

Considering the three tables above, the most common performance metrics are – Accuracy and F1-Score and MAE.

## 2.7 Fish growth models

In aquaculture, fish growth models are crucial (Kim et al., 2017). Several factors can affect shrimp growth, including water quality, feed quality and quantity, and stocking density (Tian et al., 1993), so it is essential to develop these models in order to predict fisheries outcomes, improve fish farms, and analyze the dynamics of fish populations (Kim et al., 2017). Also, stock assessments require representative growth models to be selected (Chang et al., 2022). To describe fish growth, several models have been widely used, including the von Bertalanffy growth model (VBGM) , the Gompertz growth model (Kirkwood, 2015) (Gompertz, 1825), and the logistic model . Von Bertalanffy's growth model (VBGM) is the most studied and commonly applied of all length models (Katsanevakis et al., 2008), as a result of its simplicity as well as its ability to interpret parameters biologically (Baer et al., 2011). Other models, however, better describe growth for many aquatic species (Katsanevakis et al., 2008). Growth curves provide information on how to improve rearing. Exact timing can be determined for increased growth (Baer et al., 2011).

There are 11 growth stages between the hatching of the eggs and the post-larval (PL) stage (Table 4*)*. Along with the change in length and age, there are physical changes in larvae that are mainly expressed in the eyes, rostrum, antennal flagellum, uropod, telson, pleopods, and pereopods. Distinguishing these changes distinguishes between the different zoea stages (Uno, 1969).
The data in the table below, represent larvae of the largest freshwater prawn, Macrobrachium Rosenberg. The prawn was reared from egg to post larva under the conditions of $28 \pm 0.5℃$ in water temperature, 6.58- 6.81% CL in salinity, and feeding on Artemia salina nauplii in a filter-circulative aquarium. It metamorphoses to post larva through eleven zoea stages. It should be noted that the number of days is not fixed, and larvae can reach growth stage 11 between 16 and over 30 days (Uno, 1969).

Table 4- Larvae growth stages (Uno, 1969)

| Larva stage | Age (days) | | Length (mm) | |
|---|---|---|---|---|
| | Mean | Rang | Mean | Standard deviation |
| **1** | 0 | 1-2 | 1.92 | 0.02 |
| **2** | 2 | 2-3 | 1.99 | 0.06 |
| **3** | 4 | 3-5 | 2.14 | 0.05 |
| **4** | 7 | 5-9 | 2.50 | 0.08 |
| **5** | 10 | 9-12 | 2.84 | 0.07 |
| **6** | 14 | 12-18 | 3.75 | 0.37 |
| **7** | 17 | 15-20 | 4.06 | 0.15 |
| **8** | 10 | 18-22 | 4.68 | 0.2 |
| **9** | 24 | 21-29 | 6.07 | 0.29 |
| **10** | 28 | 25-34 | 7.05 | 0.52 |
| **11** | 31 | 28-37 | 7.73 | 0.81 |
| **PL.** | 36- | 33-43 | 7.69 | 0.65 |

# 3 Methods

Two computer vision systems for aquaculture are developed in this thesis focusing on detecting and counting crustacean's larvae in industrial ponds at all larvae's growth stages (chapter 4) and crustacean's larvae outside the industrial ponds for the first larvae's growth stage (chapter 5). In addition, the ability to predict the larvae growth function based on machine vision is presented in chapter 6.

## 3.1 Counting crustacean's larvae in industrial ponds at all larvae's growth stages

A machine vision system to count crustacean's larvae in industrial ponds in controlled lab conditions was developed.

All the images were acquired in controlled illumination conditions in an industrial pond of 100 liters. The research included designing the system and developing algorithms for counting the larvae. An RGB image obtained from an iPhone 11 camera and SONY DSC-HX90V camera was used as input. Each image included several larvae (5-100).

An algorithm was developed to calculate the number of larvae in each image. Two experiments were performed in which a total of 750 images were acquired for the algorithm development along different growth stages. More details are provided in chapter 4.

## 3.2 Counting crustacean's larvae outside the industrial ponds only at day one

A machine vision system was developed to count larvae at day one (when the female lays her eggs). To ensure accurate results which are important for this stage counting was performed by manually removing larvae from the initial growing ponds with a pump and placed in a petri dish 90 millimetres in diameter under the camera. The images were acquired in controlled illumination conditions outside of the industrial ponds.

RGB images were acquired with a DSLR Nikon D510 camera using an AF-S NIKKOR 70-200mm 1:2.5G macro lens. The images were used as an input, and the algorithm output was the larvae's number in each image. Seven experiments were performed in which a total of 700 images were acquired for the algorithm development. More details are provided in chapter 5.

## 3.3 Growth function

It takes 11 growth stages for larvae to become PL from the moment they hatch. Ten larvae were measured at each stage, all from the same container. A Nikon ECLIPSE Ci-S microscope with color camera Nikon DS-Fi3 was used to capture 2048X2880 resolution images. Afterward, a function describing growth data was fitted. More details are provided in chapter 6.

## 3.4 Algorithms

Two object detection models were developed to count the larvae in the images based on 'EfficientDet D4' and 'YOLOv5s'. They are both a one stage detector algorithm. They were selected since they are considered as state-of-the-art networks object detection models. The counting algorithm receives an image as input, detects the objects and outputs the number of objects in a single image. Due to a lack of images to train the network with initial random weights, fine tuning was performed to pre-trained weights on the Common Objects in Context (COCO) dataset.

### EfficientDet

As part of the development, the 'EfficientDet D4' object detection algorithm (Figure 2) was chosen. EfficientDet is an object detection algorithm based on the convolutional neural network (CNN) (Tan et al., 2019). EfficientDet is a one stage detector design network, with its backbone based on the EfficientNet network (Tan et al., 2019). The EfficientNet backbone uses the key idea of compound scaling- scale all three dimensions, depth (number of layers), width (number of channels) or image resolution (image size), while maintaining a balance between all dimensions of the network. In addition, EfficientDet uses a multi-scale feature fusion called BiFPN.

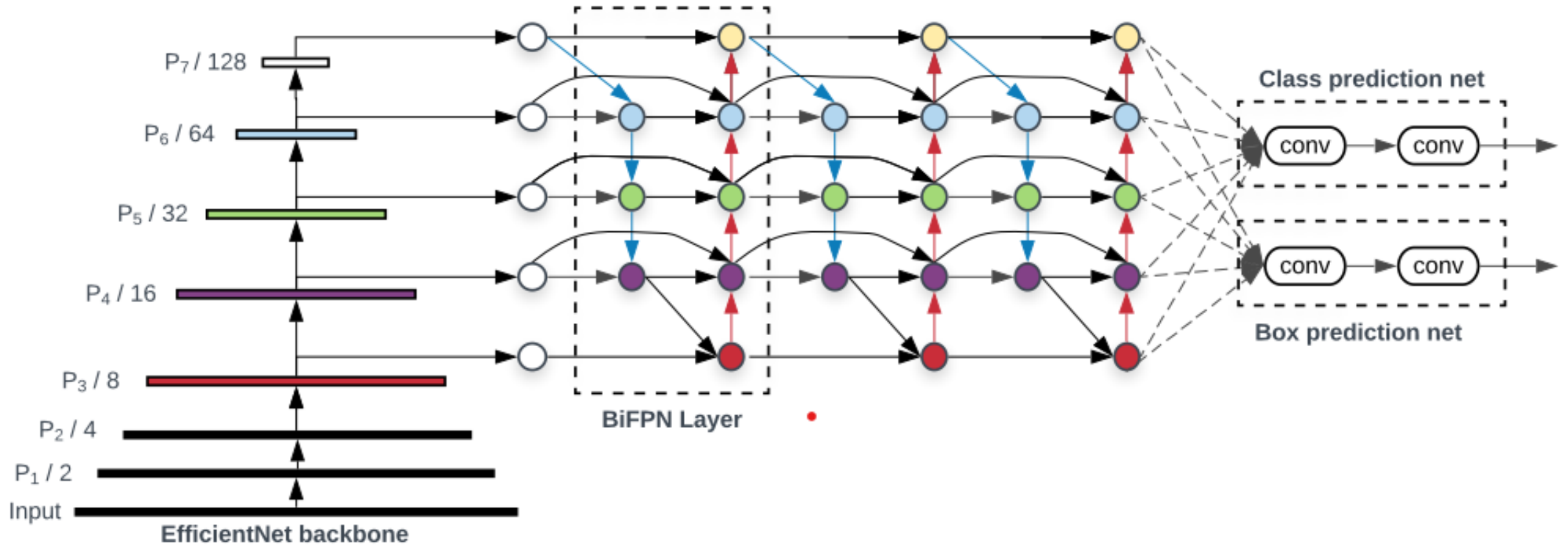

Figure 2- EfficientDet architecture

- **Cropping the images**

As can see in the table below (Table 4), in the 9-10 growth stage the larva's length is approximately 7 mm, a very small object. After running the algorithm several times, it turned out that it is necessary to crop the images to obtain a higher accuracy. The images were cropped into 2100 x 2100 resolution, so that only the bottom of the bucket (where the larvae were tagged) were left in each image, as shown in (Figure 3).

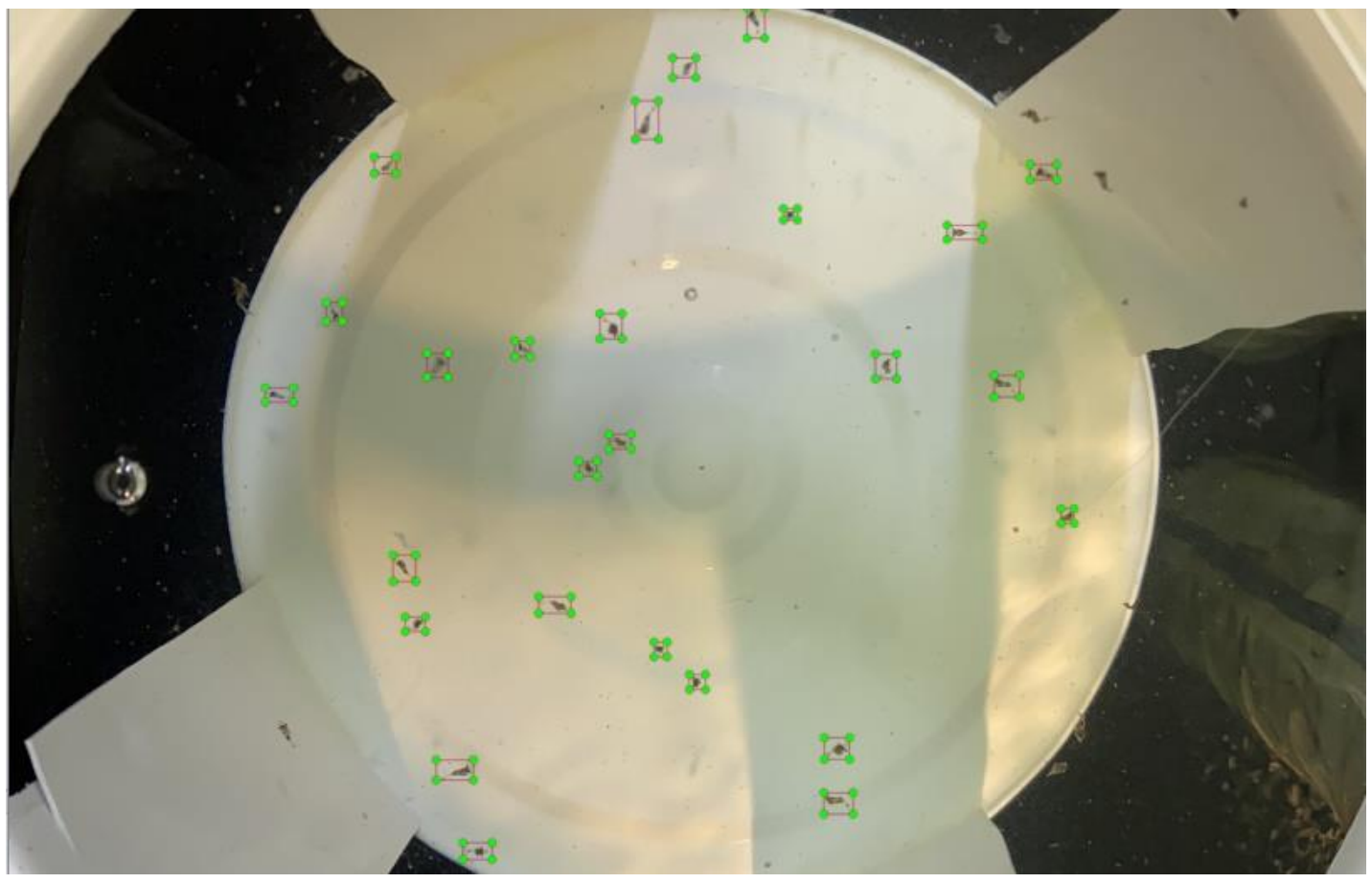

Figure 3- Cropped images

- **Resize the bounding boxes**

The algorithm could not identify the smallest objects, so it was decided to increase the bounding boxes size. To maintain the proportions of each bounding box, those that were less than the 25th percentile in relation to the area size of the entire bounding box in all the images (Figure 4 *!שגיאה מקור ההפניה לא נמצא.*), were increased.

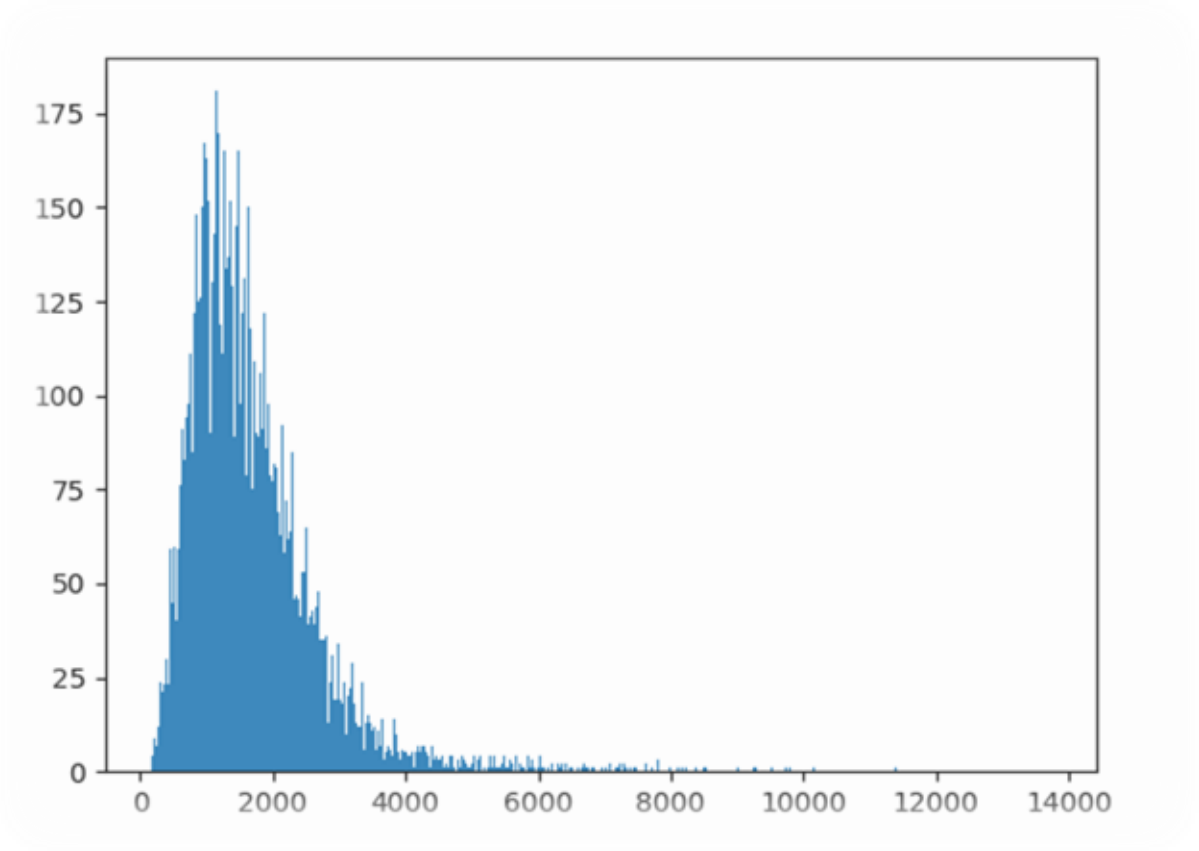

Figure 4- Plot of all bonding boxes area

The high and the wide of those bounding boxes were multiplied by an increase factor as shown in the following equation:

$$threshold\ area = The\ 25th\ percentile\ of\ all\ bbox\ areas,\ \ increase\ factor = \frac{treshold\ area}{bbox\ area}$$

**YOLOv5s**

Our algorithm is based on YOLOv5, an object detection algorithm based on the CNN convolutional neural network written in the Pytorch framework. YOLO stands for "You Only Look Once" and is one of the most popular and versatile object detection models (Plastiras et al., 2018). We selected YOLOv5 since it is much faster than the previous versions of YOLO (Jocher et al., 2020). The network backbone is CSPDarknet with PANet as the network neck. The neck is used to generate the feature pyramid network so that it can perform feature aggregation and pass it to the head for prediction (Jocher et al., 2020). YOLOv5 Head- Layers that generate predictions from the anchor boxes for object detection.

In this research the YOLOv5s was chosen because of the need to get results as quickly as possible (Jocher et al., 2020).

## 3.5 Performance measures

For object detection the concept of Intersection over Union (IoU) is used. IoU is a way to measure if a predicted bounding box is well-located, it computes the intersection over union of two bounding boxes- one for the ground truth and one for the predicted bounding box (Figure 5.שגיאה! מקור ההפניה לא נמצא.). A classification is considered as true if it matches the ground truth with IoU > N. N is the threshold for IoU, it ranges from 0 to 1. A predicted bounding box that has a big overlap with a ground truth bounding box, will have a high value of IoU.

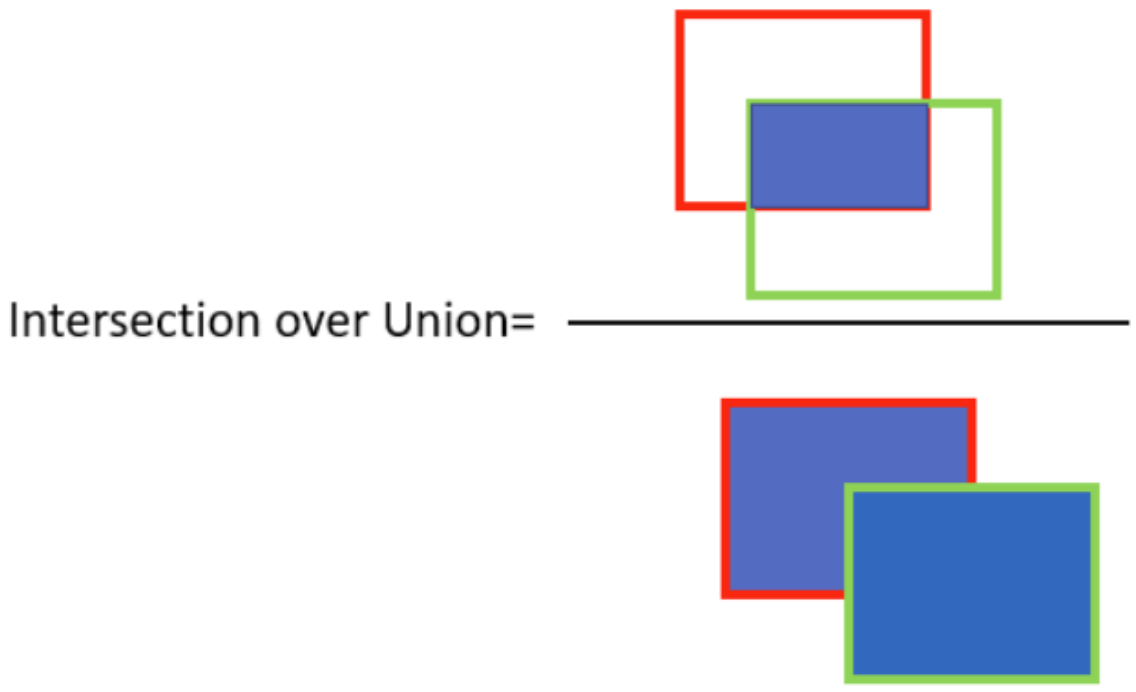

Figure 5- IOU, green- predicted b- box, red- truth b-box

The algorithm performance was evaluated with the mAP metric, combining recall and precision indicators. Given an image, an object count can be classified into one of four categories as shown in the confusion matrix (Figure 6).

| | | ACTUAL | |
|---|---|---|---|
| | | Negative | Positive |
| PREDICTION | Negative | TRUE NEGATIVE | FALSE NEGATIVE |
| | Positive | FALSE POSITIVE | TRUE POSITIVE |

Figure 6- Confusion matrix

**True Positive (TP)** - An object is a larva, and it was counted. It was counted if the ground truth and predicted bounding boxes had an IOU > N.

**True Negative (TN)** - An object is not a larva, and it was not counted. It is not useful for object detection.

**False Positive (FP)** -An object is not a larva, but it was counted.

**False Negative (FN)** -An object is a larva, and it was not counted.

With the help of TP, TN, FN, and FP, the following performance metrics can be calculated:

- The overall **accuracy** is defined as the probability of correctly classifying a test sample:

$$Accuracy = \frac{TP + TN}{Total\ number\ of\ instances}$$

- **Precision** is defined as out of all the positive predicted, what percentage is truly positive:

$$Precision = \frac{TP}{TP + FP}$$

- **Recall** is defined as out of the total positive, what percentage are predicted positive:

$$Recall = \frac{TP}{TP + FN}$$

- F1 Score is the harmonic mean of precision and recall:

$$F1\ score = 2 * \frac{Precision * Recall}{Precision + Recal}$$

The Mean Average Precision (mAP) is the area under Precision–Recall curve. The value of mAP indicates the algorithm performance.

## 3.6 Growth stage

A growth function was developed to predict the average larvae size along a growth period of 19 days. 10 larvae from each growth stage were tested in controlled indoor illumination conditions, under laboratory conditions. Total of 290 images were acquired along three weeks from using microscope -Nikon ECLIPSE Ci-S with color camera Nikon DS-Fi3 camera. The average size for each larva stage was calculated. Chapter 6 contains more details.

# 4 Counting larvae in industrial ponds

This chapter presents the development of a machine vision system for detecting and counting larvae using color images acquired from both an iPhone 11 camera and a SONY DSC-HX90V camera is described. Evaluation of the experiments conducted and results from this study are detailed.

## 4.1 System design

A system using a RGBD camera is supposed to replace manual counting of larvae with an algorithm based on an algorithm that detects and counts them. As the system needs to move between the tanks, a tripod placed on a raised surface with wheels was used to hold the camera. The camera was positioned at a 90-degree angle to allow direct observation of the larvae.

Images were acquired with two different cameras: an iPhone 11's camera (resolution of 3024X4032), *CAMERA Specs- Appendix E* and a SONY DSC-HX90V camera (resolution of 1896X3672), *CAMERA Specs- Appendix F*.

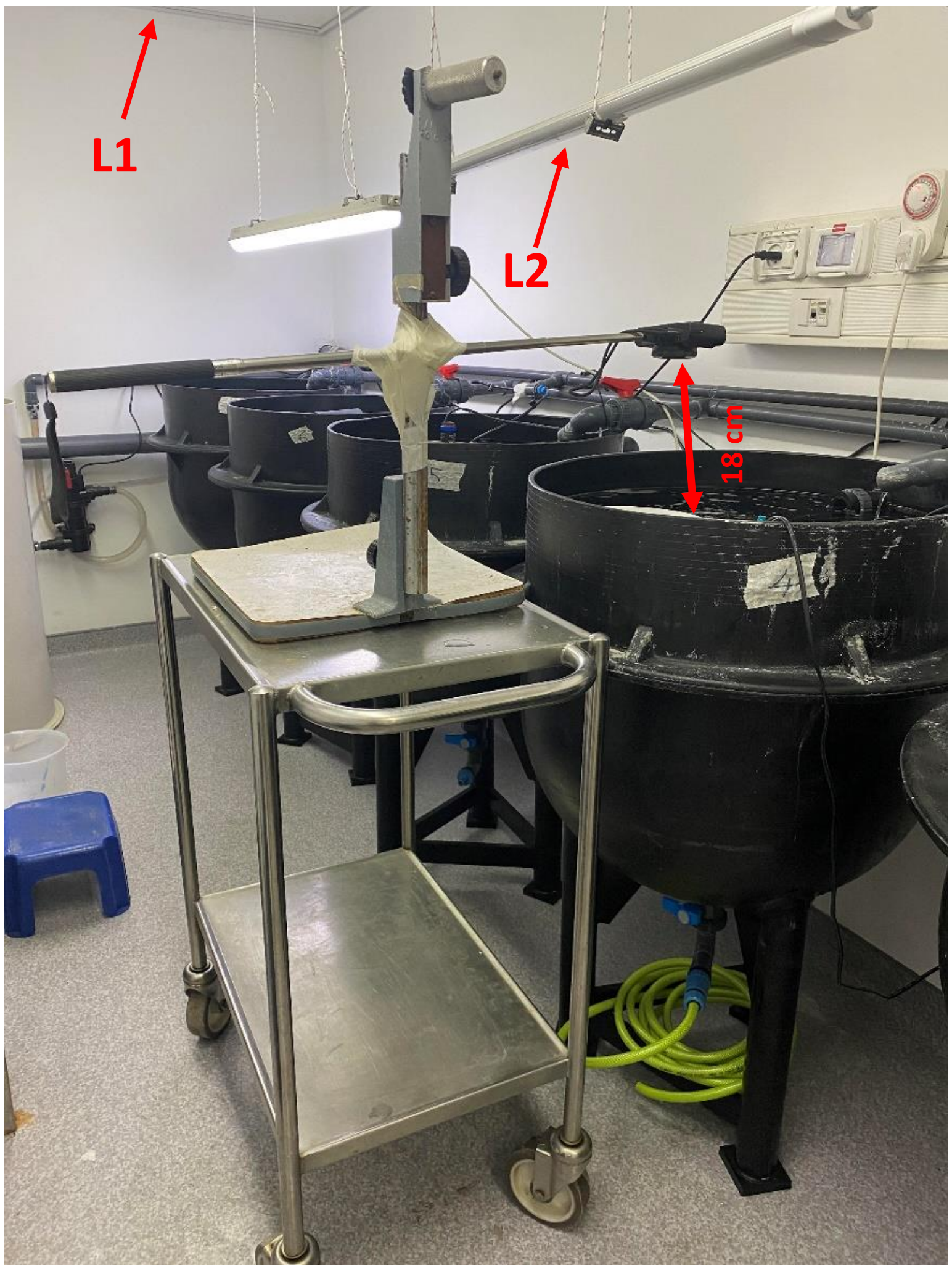

Figure 7- System design. Camera's distance was 18 cm from the pond's top

## 4.2 Experimental design

The experiment was performed in larvae industrial ponds in controlled at Ben-Gurion University (Figure 8). The size of each pond is 600 mm diameter, and 600 mm depth – resulting in approximately 100 liters. The growing ponds have water filters. The room is dark and above the ponds there is illumination (LEDs for white lighting ip65 FLUX 3200lm) to maintain controlled conditions.

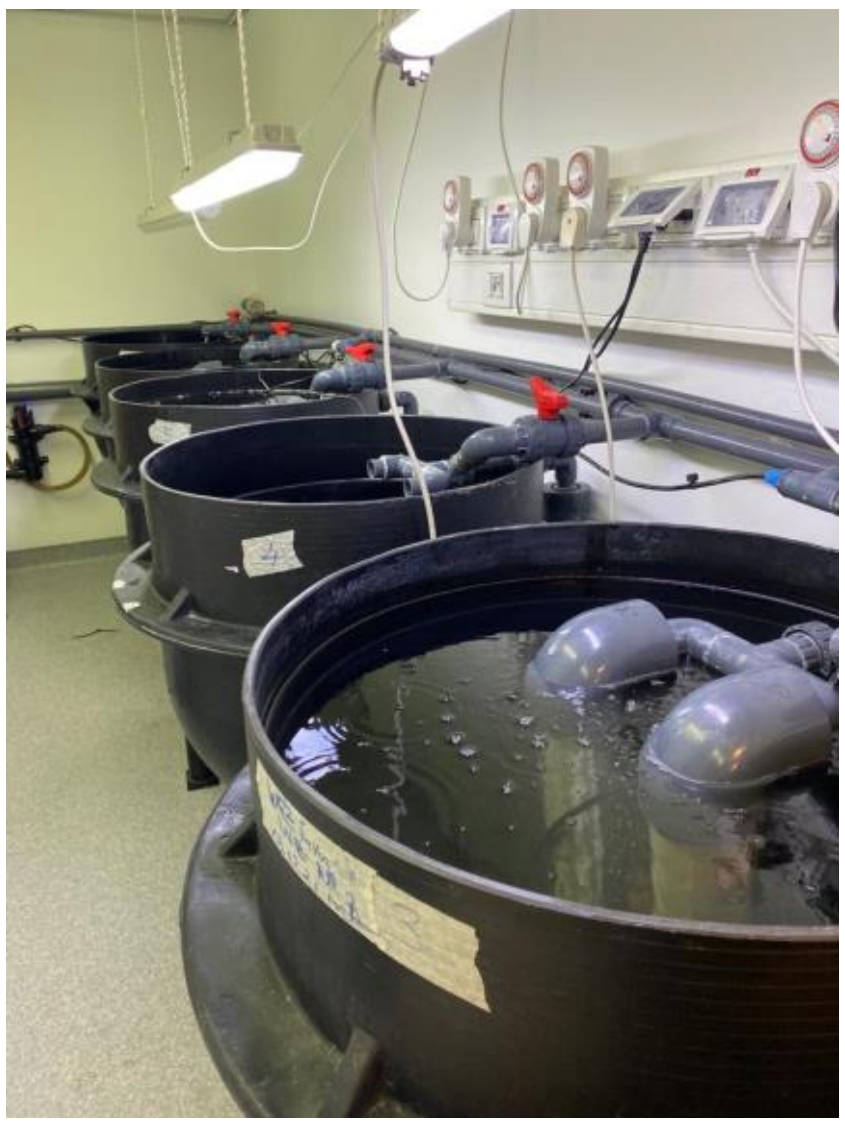

Figure 8- Larvae growing ponds at BGU

Before taking the pictures, the water was mixed to create a uniform distribution of the larvae in the pond. The ponds themselves are tightly painted and the color of the larvae is relatively dark, so they are not prominent enough in the picture. To overcome this, a white bucket ( שגיאה! מקור ההפניה לא נמצא.שגיאה! מקור ההפניה לא נמצא.) was placed in the pond, the size of the bucket is 20 cm diameter and 19 cm height.  To interfere as little as possible with the movement of the water and the larvae's normal living conditions, the bucket was cut on its sides.

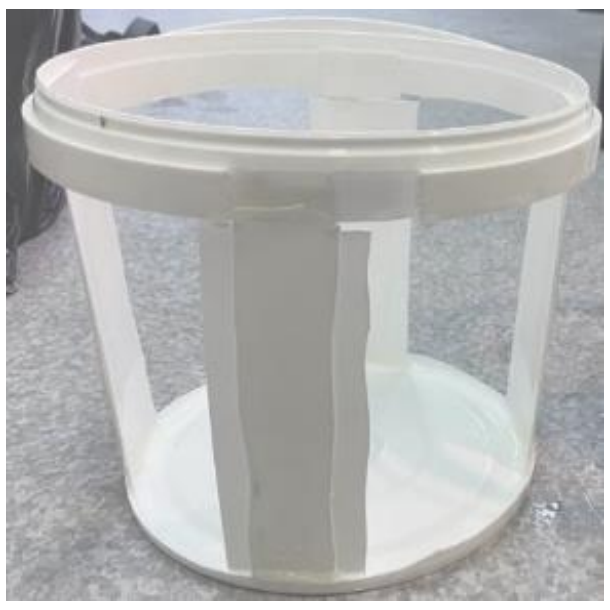

Figure 9- The bucket

## 4.3 First experiment

### 4.3.1 Data

200 images were acquired with the iPhone 11 camera. The images were acquired once on a single day during 9-10 larvae's growth stage of total 11 growth stages. The size of the photographed water volume was 20 cm diameter and 19 cm height, rounded to 6 liters. Therefore, the volume of the photographed area is about 6% of the total water's volume. The number of larvae in a pond was calculated by multiplying the number of larvae in a single image by 16.6.

After acquiring the images, they were manually labelled (Figure 10Results **שגיאה! מקור ההפניה לא נמצא.**) with 'LabelImg', a graphical image annotation tool. The dataset includes images in which bounding boxes around every larva at the bottom of the bucket only, with the class tag –'Larvae' marked.

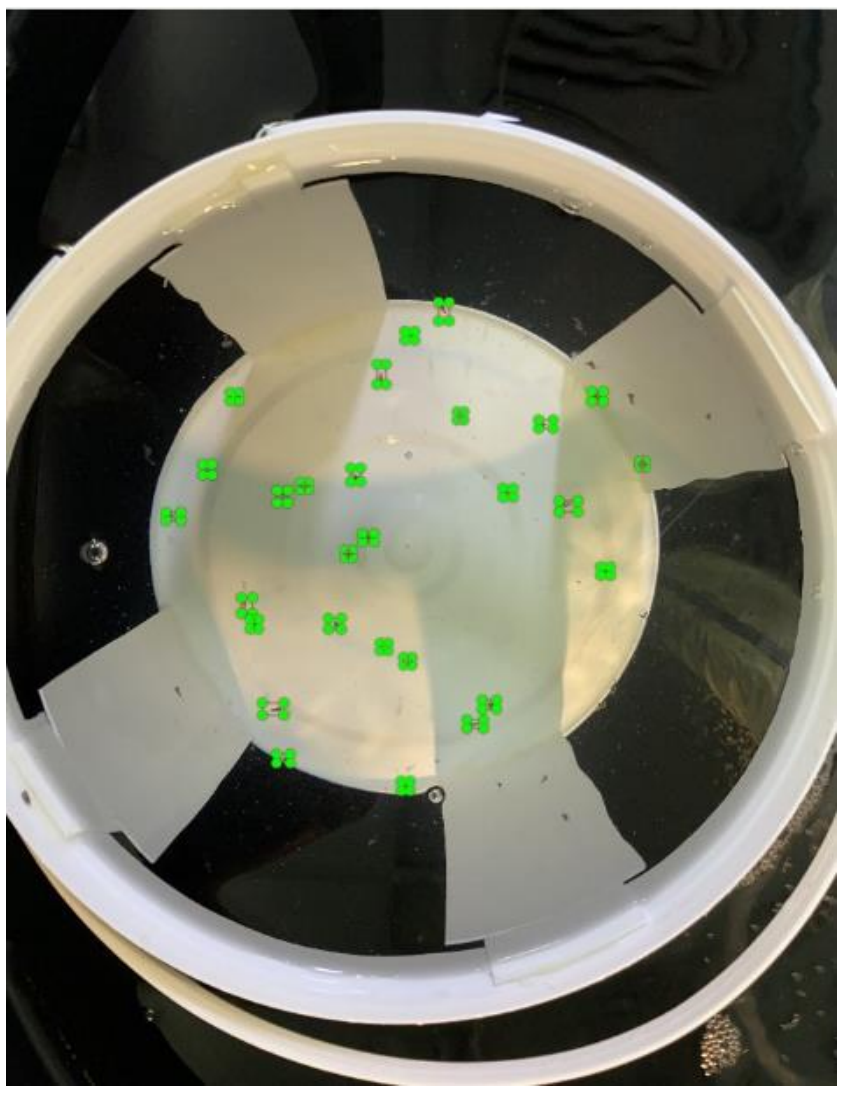

Figure 10- Labelled image

### 4.3.2 Results

The IoU was selected to be 0.5, implying 50% overlap between the ground truth and the prediction bounding box considered as a correct prediction. Results (Table5 ) reveal the best result for EfficientDet (image size- 2100, epochs- 300, batches- 16), and YOLOv5s are with the default values (image size- 640, epochs-300 , batches-16 ). The mAP YOLOv5s results were better by 31.33%, with better accuracy results by 3.89%, so it was used for the next stage.

Table5 - Preliminary results

| | EfficientDet | YOLOv5s |
|---|---|---|
| mAP | 0.57 | 0.83 |
| Accuracy | 84% | 87.4% |

It is important to note that YOLOv5s also yields faster performance than EfficientDet (Figure 11).

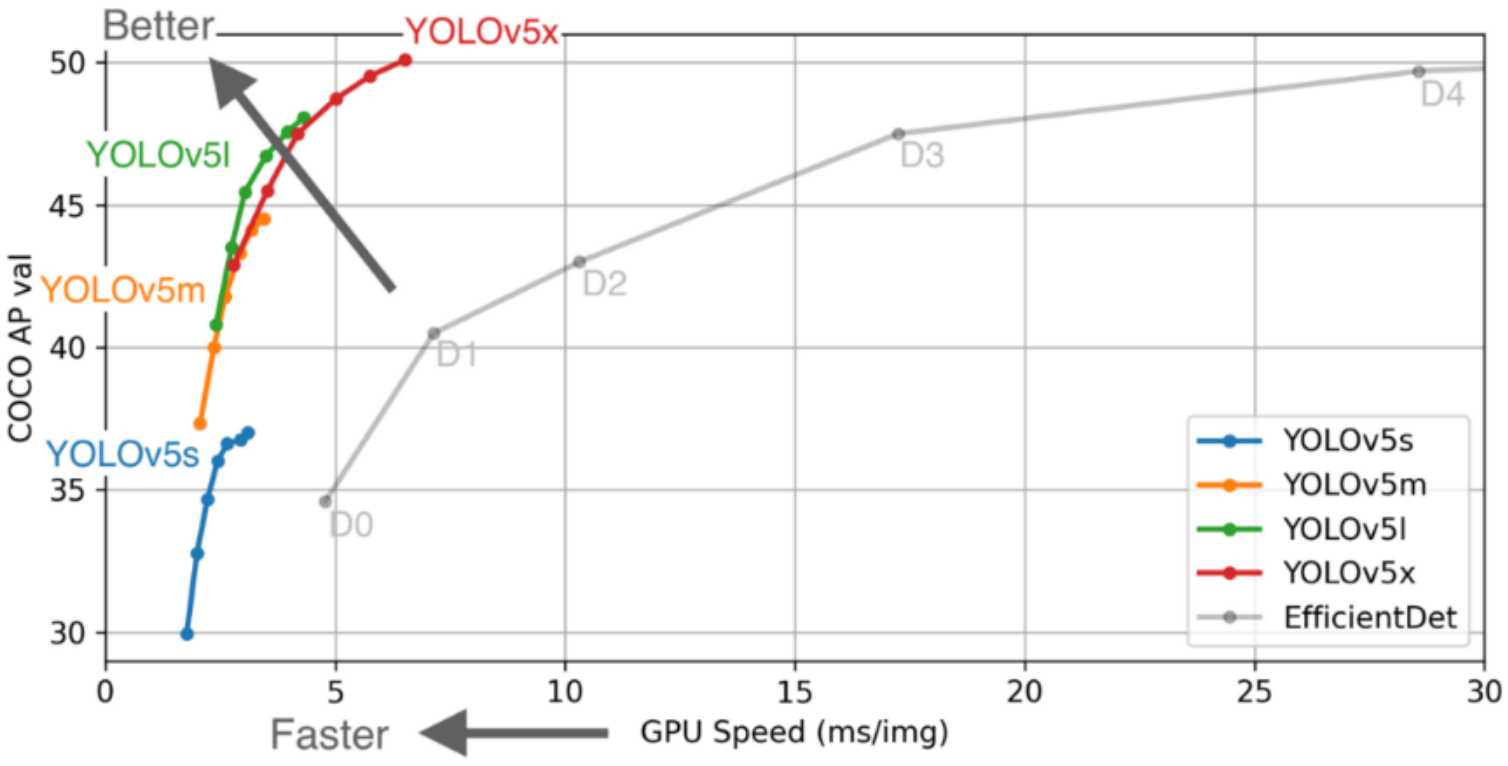

Figure 11- Detection accuracy and computational power

**Using mask**

Since the area of interest in the images is only the bottom of the bucket- the larvae were labeled only there, a circle mask was applied to the images. The mask turns any pixel outside of the bucket radius black. All images in the training, validation, and test set were masked.

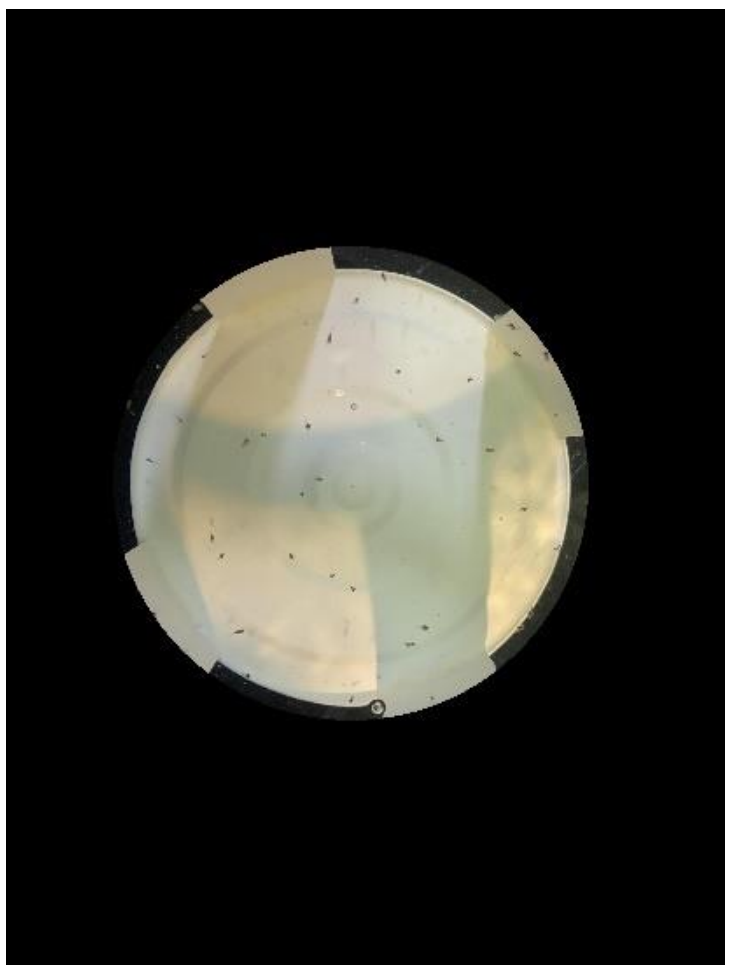
Figure 12- Image with mask

The object detection algorithm was trained and tested using 200-tagged images, 160 images in the training set and 40 images in the test set.

The final step is to perform a hyperparameters tuning to the chosen network – YOLOv5s. The image size, number of batches and the number of epochs were changed. The image size in YOLOv5s algorithm must be in multiples of 32. The number of epochs is a hyperparameter that controls the number of complete passes through the training dataset. The number of batches is the number of training samples to work through before the model's internal parameters are updated. It is customary to choose batches as numbers that are power of 2. The GPU used in the work (CUDA:0 (Tesla P100-PCIE-16GB, 16281MiB)) had a batch limit for a certain image size.

As can be seen in the table above (Appendix G - Hyper parameters tuning), as the image size increases the object's size also increases. Due to the fact the objects are very small, this helps the algorithm recognize the objects better. Nonetheless, at a size of 1536 pixels, an over-fitting was observed since the mAP index in the training set was higher than the index in the test set in every run. In addition, the higher the number of epochs, the higher the accuracy and mAP as expected However, this increases the training time. Considering both measures (Accuracy and mAP on test set) the best results (Table 6) were for a model developed with 400 epochs, image size of 1504 and 16 batches.

Table 6- Best model

| Image size | Epochs | Batch | mAP-train set | mAP-test set | Accuracy (%) |
|---|---|---|---|---|---|
| 1504 | 400 | 16 | 0.963 | 0.961 | 97 |

mAP, Precision and Recall results on validation set, where x axis represents number of epochs and y axis represent the index value.

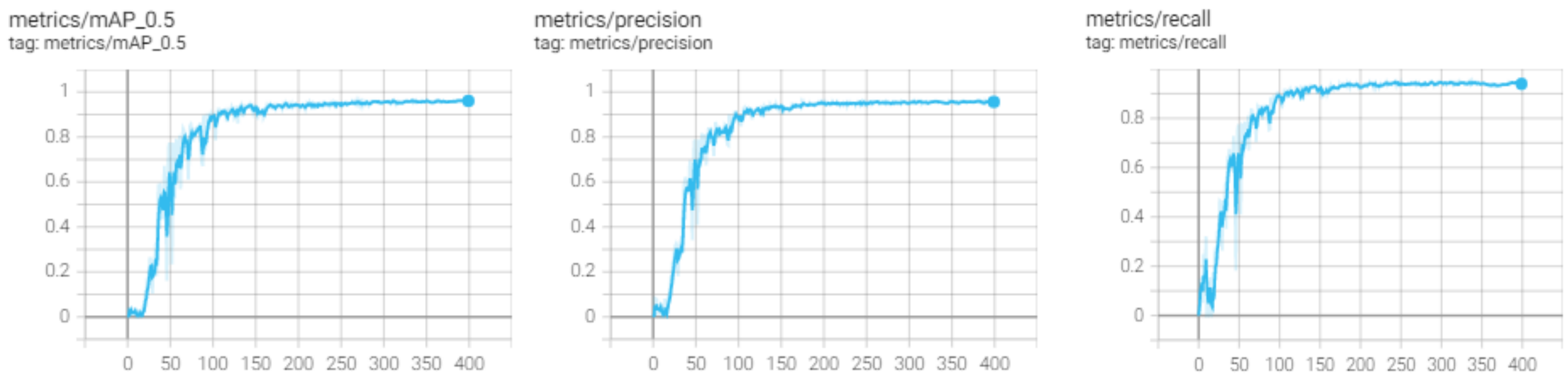

Figure13 - From left to right: mAP, Precision and Recall results on validation set

'Box loss' results on train and validation set, where the x axis represents number of epochs, and the y axis represents the loss value. The 'box loss' represents how well the algorithm can locate the center of an object and how well the predicted bounding box covers an object. The loss decreases while training, indicating the convergence of the model.

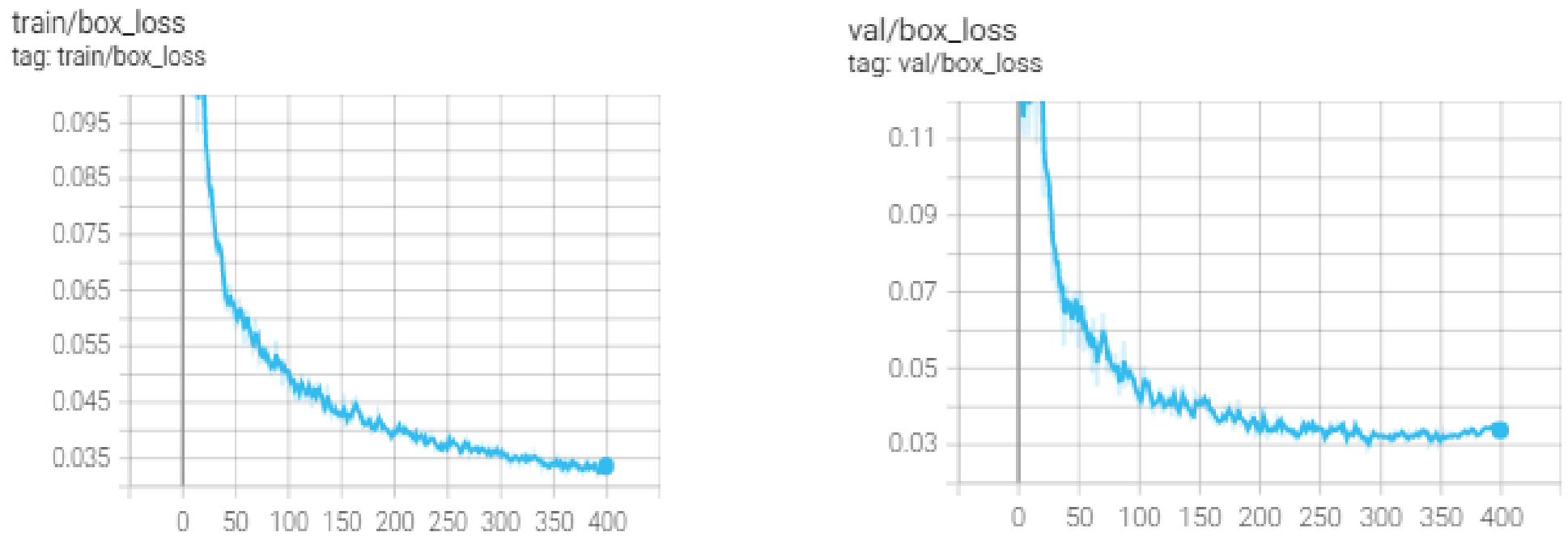

Figure14 - From left to right: box loss on train set and validation set.

The confusion matrix (Table 7) for IoU=0.5, revealed 92.875% true positive.

Table 7- Confusion matrix

| | **Predicted Positive** | **Predicted Negative** |
|---|---|---|
| **Positive** | 1851 | 72 |
| **Negative** | 70 | 0 |

The table below presents the Precision, Recall and F1-Score values, which were calculated based on confusion matrix values.

Table 8- Precision, Recall and F1-Score

| | **Precision** | **Recall** | **F1-Score** |
|---|---|---|---|
| IoU= 0.5 | 0.9635 | 0.9625 | 0.963 |

### 4.3.3 Sensitivity analysis

**Train - test size**

The impact of different number of images in the training and test sets was evaluated (Figure 15) for different train test splits (80-20%, 70-30%, 60-40%).

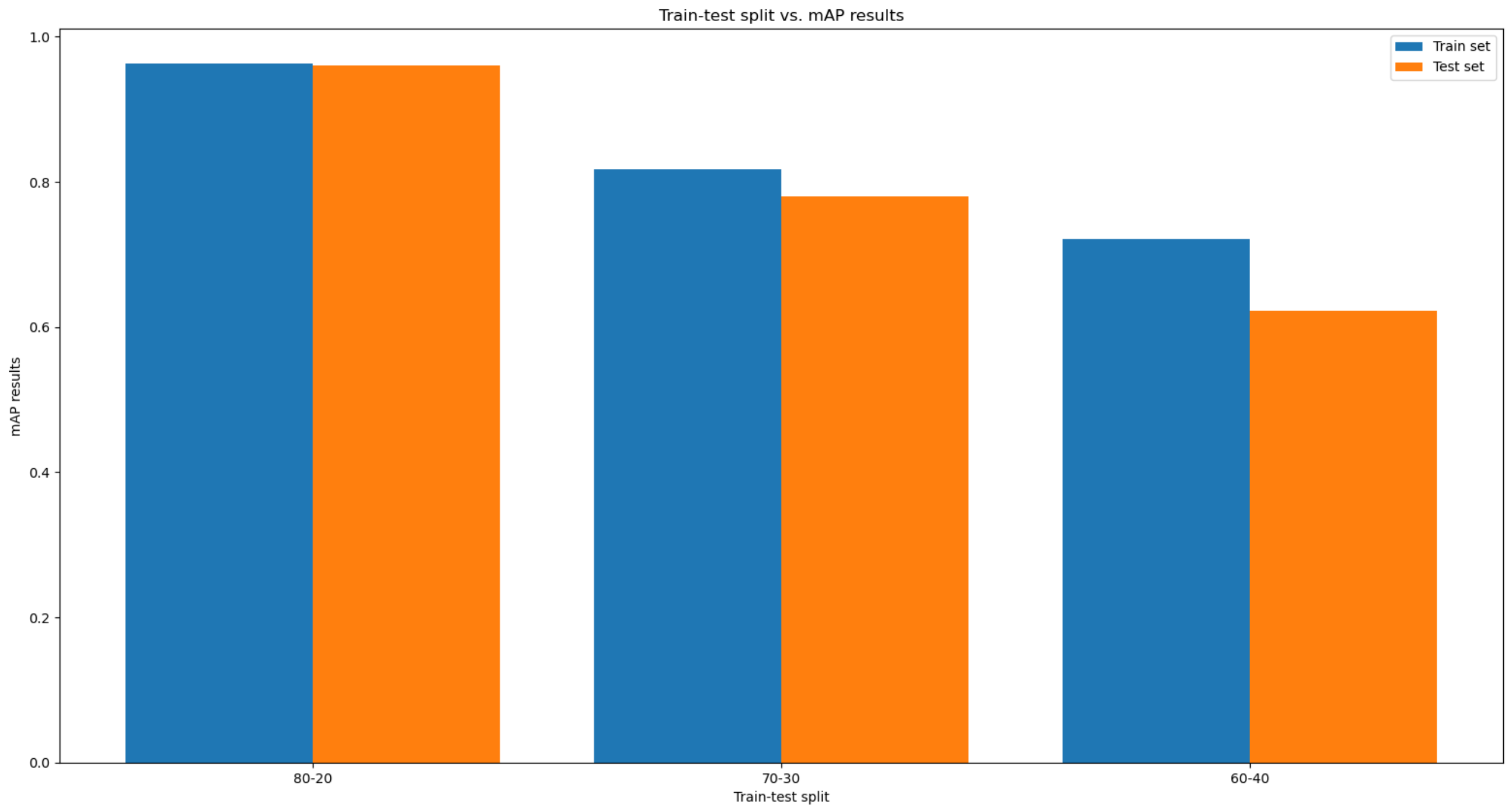

Figure 15- Different train-test split mAP results

The best performance was achieved in the first group for all indices. Moreover, results of the second and third group revealed over-fitting, the model fit the training set too well and failed on the mAP test set performance.  Also, accuracy, precision, recall and F1-Score performance are the best for 80-20% split as expected with performance decreasing as the number of images in the train decreases.

Table 9- Different train-test split Accuracy, Precision, Recall and F1-Score results

| | Accuracy | Precision | Recall | F1-Score |
|---|---|---|---|---|
| 80%-20% | 97 | 0.871 | 0.856 | 0.887 |
| 70%-30% | 85.6 | 0.76 | 0.744 | 0.777 |
| 60%-40% | 74.4 | 0.939 | 0.938 | 0.94 |

**Noisy images**

The quality of digital images is affected by different types of noise. Noisy images are those contaminated by unwanted information (Mohammed Abd-Alsalam et al., 2016). The Gaussian noise was tested as the most common natural noise that affects images. We evaluated performance for different noise values. Each time a noise with a different variance was added to each pixel in the test set images, with values between 5-25. The average noise was always 0. The model was trained and validated with the same test of images randomly selected. The training set contained 128 images and the test set 40 images (80-20% split). As the noise variance increases, the accuracy values of the model decreases (Table 10), as expected, but the model still manages to detect the larvae very well.

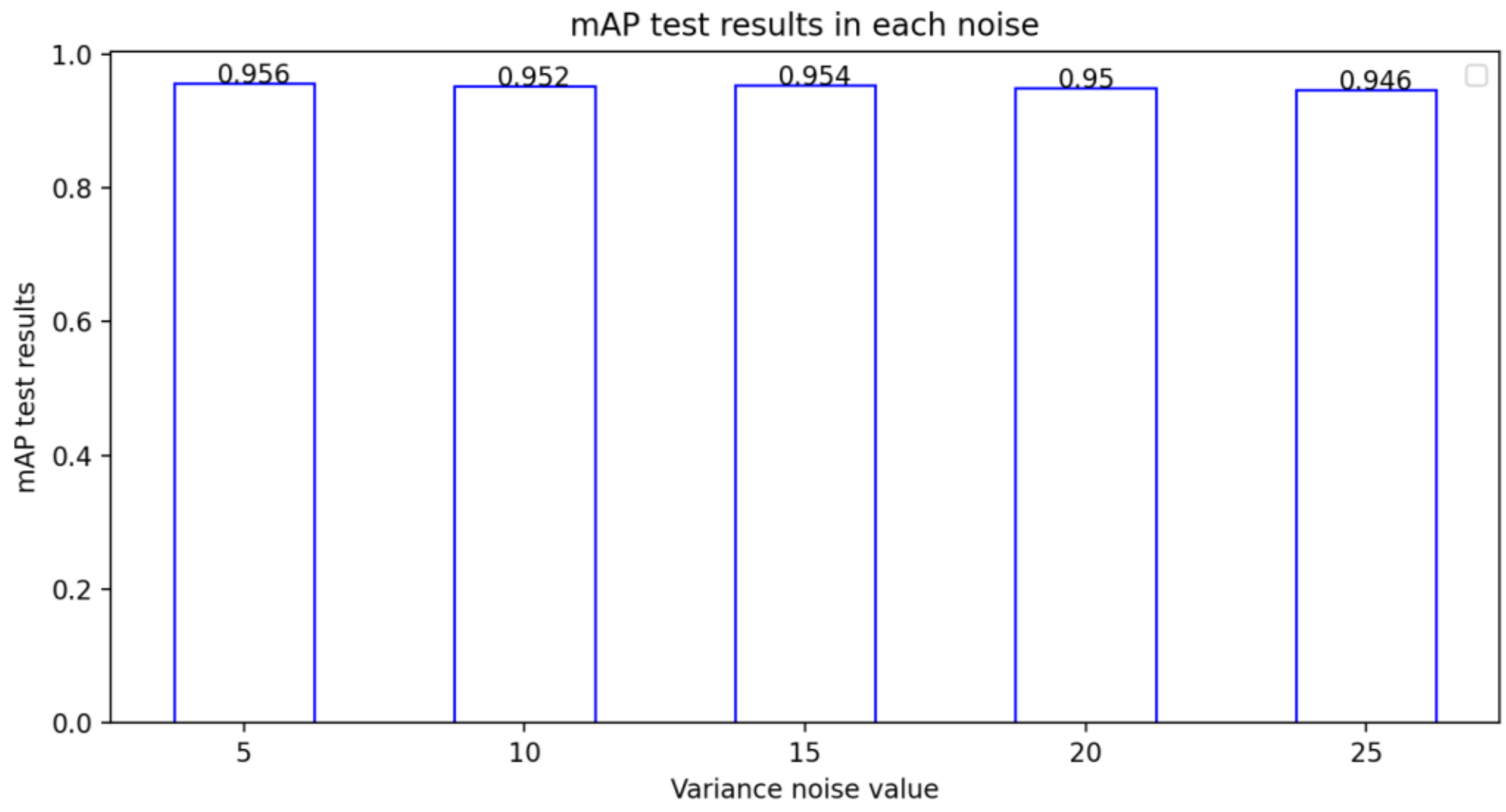

Figure 16- Different noise mAP results

Table 10- Different noise variance values Accuracy, Precision, Recall and F1-Score results

| Noise variance | Accuracy | Precision | Recall | F1-Score |
|---|---|---|---|---|
| 5 | 96.5 | 0.9636 | 0.9625 | 0.9657 |
| 10 | 96.2 | 0.9625 | 0.962 | 0.9623 |
| 15 | 96.4 | 0.9626 | 0.9634 | 0.9632 |
| 20 | 96 | 0.962 | 0.9605 | 0.9612 |
| 25 | 95.7 | 0.9589 | 0.9573 | 0.958 |

## 4.4 Second experiment

### 4.4.1 Data

In the second experiment, a larvae industrial pond was photographed along 11 days, starting from a few days after the hatching of the eggs until they reached their post-larval stage. In this experiment two devices were used for image acquisition: an iPhone 11 camera and a SONY DSC-HX90V camera. With the first device (iPhone 11) two illumination conditions were tested. In the first condition (L1), the lamps above the growth pond were off, but the room lighting was kept on. In the second illumination condition (L2) the lamps above the growth pond were turned on, but the room lighting was kept off (Figure 7). The second illumination condition is the natural scenario. In each condition, 110 images with a resolution of 3024X4032 were acquired. Like the first experiment before taking the pictures, the water was mixed, and a white bucket (**שגיאה! מקור ההפניה לא נמצא.**) was placed in the pond. The photographed water was 20 cm diameter and 19 cm height, resulting in 6 liters.

### 4.4.2 Results

**Weights obtained from first experiment**

The network weights were set as the weights obtained from running the images of the first experiment with the best parameters- image sizes of 1504, 150 epochs, and 16 batches. No further training was conducted. The results are presented for the test sets which included data in the different acquisition conditions (cameras, illumination, and growth stages). The metrics tested in each run are mAP, TP, FP (in percentages) as described in chapter 3.5.

- ***images without a mask, neither in the training set nor in the test set***

Table 11- Weights obtained from first experimental results of SONY camera- L2 and iPhone 11 camera- L2, L1

| | *SONY camera* | | | *iPhone 11 camera* | | | | | |
|---|---|---|---|---|---|---|---|---|---|
| | **L2** | | | **L1** | | | **L2** | | |
| ***Date*** | **images** | **Accuracy (%)** | **mAP** | **images** | **Accuracy (%)** | **mAP** | **images** | **Accuracy (%)** | **mAP** |
| *18/10/2021* | 10 | 3.4 | 0.002 | 10 | 36.9 | 0.239 | 10 | 1.3 | 0.001 |
| *19/10/2021* | 10 | 21.8 | 0.027 | 10 | 348 | 0.193 | 10 | 68.4 | 0.472 |
| *20/10/2021* | 10 | 48.9 | 0.145 | 10 | 32.9 | 0.113 | 10 | 79.6 | 0.641 |
| *21/10/2021* | 10 | 50.7 | 0.192 | 10 | 20.6 | 0.151 | 10 | 71.2 | 0.646 |
| *24/10/2021* | 10 | 83 | 0.42 | 10 | 76.2 | 0.707 | 10 | 82.5 | 0.672 |
| *25/10/2021* | 10 | 71.4 | 0.447 | 10 | 100 | 0.735 | 10 | 68.1 | 0.531 |
| *26/10/2021* | 10 | 84.8 | 0.624 | 10 | 100 | 0.98 | 10 | 71.7 | 0.632 |
| *27/10/2021* | 10 | 91 | 0.727 | 10 | 83.6 | 0.641 | 10 | 92.8 | 0.812 |
| *29/10/2021* | 10 | 82.2 | 0.597 | 10 | 89.2 | 0.618 | 10 | 90.3 | 0.641 |
| *31/10/2021* | 10 | 91.4 | 0.803 | 10 | 68.6 | 0.643 | 10 | 92.1 | 0.786 |
| *03/11/2021* | 10 | 88.5 | 0.731 | 10 | 66.2 | 0.537 | 10 | 73.4 | 0.594 |
| ***Sum*** | **110** | | | **110** | | | **110** | | |
| ***Average*** | | **65.19** | **0.429** | | **64.454** | **0.505** | | **71.945** | **0.584** |

The best results were obtained for iPhone 11 camera under the second illumination conditions as shown in the table above (Table 11). Upon receiving the results, another 20 images were acquired each day with the iPhone11 camera under illumination conditions 2 (a total of 220 additional photos). The results were recalculated again, this time for 30 images on each day (Table 12).

Table 12- Weights obtained from first experimental results of iPhone 11 camera- L2

| | *iPhone 11 camera- L2* | | |
|---|---|---|---|
| ***Date*** | **images** | **Accuracy (%)** | **mAP** |
| *18/10/2021* | 30 | 7.7 | 0.018 |
| *19/10/2021* | 30 | 69.9 | 0.432 |
| *20/10/2021* | 30 | 81.9 | 0.523 |
| *21/10/2021* | 30 | 68.9 | 0.579 |
| *24/10/2021* | 30 | 80.4 | 0.619 |
| *25/10/2021* | 30 | 73.8 | 0.521 |
| *26/10/2021* | 30 | 76.7 | 0.579 |
| *27/10/2021* | 30 | 88.4 | 0.743 |
| *29/10/2021* | 30 | 89.5 | 0.692 |
| *31/10/2021* | 30 | 91 | 0.776 |
| *03/11/2021* | 30 | 82.5 | 0.689 |
| ***Average*** | | **73.691** | **0.561** |

The best results in both measures- accuracy and mAP were obtained for the date 31/10/2021. These results were better by 4.44% in mAP (obtained for the date 27/10/2021) and by 1.676% in accuracy (obtained for the date 29/10/2021) respectively than the next best results, and by 4,211% from the worst results in mAP and in accuracy 1,081% (both obtained for the date 18/10/2021).

- ***images with a mask, in the training and the test set***

Table 13- Weights obtained from first experimental results of SONY camera- L2 and iPhone 11 camera- L2, L1

| | *SONY camera* | | | *iPhone 11 camera* | | | | | |
|---|---|---|---|---|---|---|---|---|---|
| | | **L2** | | | **L1** | | | **L2** | |
| ***Date*** | **images** | **Accuracy (%)** | **mAP** | **images** | **Accuracy (%)** | **mAP** | **images** | **Accuracy (%)** | **mAP** |
| *18/10/2021* | 10 | 1.7 | 0.001 | 10 | 29.1 | 0.219 | 10 | 1.6 | 0.006 |
| *19/10/2021* | 10 | 15.8 | 0.07 | 10 | 34.9 | 0.269 | 10 | 78.4 | 0.693 |
| *20/10/2021* | 10 | 45.5 | 0.324 | 10 | 33.3 | 0.266 | 10 | 73.9 | 0.678 |
| *21/10/2021* | 10 | 37.5 | 0.275 | 10 | 19 | 0.159 | 10 | 71.6 | 0.636 |
| *24/10/2021* | 10 | 68.7 | 0.537 | 10 | 85 | 0.808 | 10 | 86.1 | 0.749 |
| *25/10/2021* | 10 | 78.6 | 0.733 | 10 | 100 | 0.806 | 10 | 82.1 | 0.774 |
| *26/10/2021* | 10 | 90.5 | 0.856 | 10 | 92.3 | 0.914 | 10 | 86.8 | 0.795 |
| *27/10/2021* | 10 | 79.2 | 0.672 | 10 | 84.5 | 0.683 | 10 | 88.8 | 0.812 |
| *29/10/2021* | 10 | 62 | 0.421 | 10 | 91.1 | 0.852 | 10 | 89.4 | 0.675 |
| *31/10/2021* | 10 | 82.3 | 0.751 | 10 | 91.3 | 0.878 | 10 | 94 | 0.8 |
| *03/11/2021* | 10 | 75.4 | 0.644 | 10 | 85.8 | 0.781 | 10 | 81.5 | 0.719 |
| ***Sum*** | **110** | | | **110** | | | **110** | | |
| ***Average*** | | **64.154** | **0.55** | | **67.845** | **0.603** | | **75.836** | **0.667** |

On both measures the best results were obtained for the iPhone 11 camera under the second illumination conditions also. The results were improved compared to the results without the mask as accepted (Table 11) - better by 14.21% in mAP and by 5.408% in accuracy.

As explain above 30 more images were tagged for the best camera and illumination results. Also, here for iphone11 camera under L2 condition.

Table 14- Weights obtained from first experimental results of iPhone 11 camera- L2

*iPhone 11 camera- L2*

| *Date* | images | Accuracy (%) | mAP |
|---|---|---|---|
| *18/10/2021* | 30 | 5.9 | 0.037 |
| *19/10/2021* | 30 | 76.7 | 0.642 |
| *20/10/2021* | 30 | 76.9 | 0.7 |
| *21/10/2021* | 30 | 67.9 | 0.568 |
| *24/10/2021* | 30 | 85.1 | 0.716 |
| *25/10/2021* | 30 | 84.3 | 0.793 |
| *26/10/2021* | 30 | 89.4 | 0.825 |
| *27/10/2021* | 30 | 86.5 | 0.763 |
| *29/10/2021* | 30 | 88.3 | 0.705 |
| *31/10/2021* | 30 | 92.2 | 0.772 |
| *03/11/2021* | 30 | 86.3 | 0.756 |
| ***Average*** | | **76.318** | **0.6615** |

Indeed, in both parameters - accuracy and mAP the data set with the mask yielded better results. The best results for iPhone 11 camera under the second illumination conditions accuracy increases by 17.9% and mAP increases by 3.565%.

**Training each illumination condition and camera separately (with mask)**

Each illumination and camera condition were trained separately with each of the three combinations including all the images from all days. A random sample of 80% of the images was used for training, and 20% for testing. Each combination was run twice - first with the data as-is and then with augmentations (90° Rotate) of the training set added.

Table15 - Each illumination condition and camera separately results. image size -1504, epochs-400, batches- 16

| *Train* | *Test and validation* | *Images (per day)* | *AUG* | *mAP- train* | *Accuracy (%)* | *mAP- test* |
|---|---|---|---|---|---|---|
| SONY-L2 | SONY-L2 | 10 | NO | 0.695 | 70.8 | 0.635 |
| SONY-L2 | SONY-L2 | 10 | YES | 0.836 | 72.5 | 0.687 |
| iPhone11-L1 | iPhone11-L1 | 10 | NO | 0.758 | 79.6 | 0.746 |
| iPhone11-L1 | iPhone11-L1 | 10 | YES | 0.823 | 80.1 | 0.771 |
| iPhone11-L2 | iPhone11-L2 | 10 | NO | 0.809 | 74.3 | 0.67 |
| iPhone11-L2 | iPhone11-L2 | 10 | YES | 0.737 | 67.2 | 0.562 |
| iPhone11-L2 | iPhone11-L2 | 30 | NO | 0.909 | 88.4 | 0.855 |

The best results were obtained for the train and test set of iPhone 11 camera under the second illumination conditions with 30 images per day, no augmentation. The result agrees with the results above for each day as the best results were obtained for the same conditions. The mAP index results were better by 10.895% over the previous best results and by 10.36% over the worst results. Applying augmentations on the images improved the results under SONY-L2 and iphone11-L1 condition, In contrast to iphone11-L1 condition.

### 4.4.3 Sensitivity analysis

**Camera effect**

A pair of runs with neutral lighting conditions was conducted to determine whether the camera type affected the results. During the first run, the training set was an iPhone11 camera in illumination conditions 2, while the validation and test set was a SONY camera in illumination conditions 2, and vice versa in the second run. The results below (Table17 ) show that using an iPhone11 camera yields more accurate training and better mAP results in both- train and test sets.

Table16 - Camera effect. image size -1504, epochs- 400, batches- 16

| *Train* | *Test and validation* | *Images (per day)* | *AUG* | *mAP- train* | *Accuracy (%)* | *mAP- test* |
|---|---|---|---|---|---|---|
| iPhone11-L2 | SONY-L2 | 10 | NO | 0.68 | 63.1 | 0.557 |
| SONY-L2 | iPhone11-L2 | 10 | NO | 0.552 | 50.3 | 0.373 |

Camera type does affect the results, the learning of the algorithm is done in the best way when the training is done on images purchased by iphon11-L2. Results improved by 49.33% in mAP and 25.45% in accuracy.

**Illumination effect**

A pair of runs with neutral camera conditions was conducted to determine whether the illumination condition affected the results. During the first run, the training set was set with the iPhone11 camera in L1, while the validation and test set was with the same camera in L2, and vice versa in the second run. The results below (Table17 ) show that using L2 yields more accurate training and better mAP results on the test set. The difference between the mAP results in the train and test sets in the first run, indicate on over-fitting.

Table17 - Illumination effect. image size -1504, epochs- 400, batches- 16

| ***Train*** | ***Test and validation*** | ***Images (per day)*** | ***AUG*** | ***mAP- train*** | ***Accuracy (%)*** | ***mAP- test*** |
|---|---|---|---|---|---|---|
| iPhone11 -L1 | iPhone11 -L2 | 10 | NO | 0.728 | 71.6 | 0.587 |
| iPhone11 -L2 | iPhone11 -L1 | 10 | NO | 0.723 | 74.1 | 0.702 |

It is best to train the algorithm on images that were purchased in the second illumination condition to get the best results. mAP and accuracy improved by 19.59% and 3.49%, respectively compared to results received from training the algorithm on images that purchased in the first illumination condition.

**Transfer learning**

Transfer learning is popular for building accurate models in a time-saving manner (Rawat & Wang 2017). Instead of starting over from scratch, transfer learning starts with patterns learned from solving similar problems. Transfer learning occurs when a model is trained on a similar problem and applied to another one (Brodzicki et al., 2020), it allows us to build accurate models in a time saving way. The transfer learning technique widely used in computer vision, takes a source task for the training and uses its results as the starting point for a different new task (Guo et al., 2019). Transfer learning based on networks can be classified into two categories- feature extraction and fine-tuning. In feature extraction a new classifier is trained based on the pre-trained base model. Only the fully connected layer is trained while the convolution layers' weights remain unchanged. The fine-tuning method involves both, retraining the fully connected layer and adjusting a few of the convolution layers as well (Brodzicki et al., 2020). Transfer learning techniques were used to train the model on one camera type and illumination condition, then retrain the convolution layers on the other. A total of 10 images per day were trained, total of 22 images were evaluated using transfer learning methods with the same hyperparameters (image size of 1504, 400 epochs and 16 batches).

- Initial learning rate remains 0.01:

**Feature extraction- freeze 24 convolution layers**

Table18 - Transfer learning. Initial learning rate of 0.01, feature extraction

| ***Pretrain set*** | | ***Retrain set*** | | ***mAP- train*** | ***Accuracy (%)*** | ***mAP- test*** | ***Converge epoch*** | ***Last training epoch*** | ***Time (hours) of training*** |
|---|---|---|---|---|---|---|---|---|---|
| *Camera type* | *Illumination condition* | *Camera type* | *Illumination condition* | | | | | | |
| *iPhone 11* | *1* | *iPhone 11* | *2* | *0.769* | *71.8* | *0.621* | *83* | *184* | *0.247* |
| *iPhone 11* | *2* | *iPhone 11* | *1* | *0.668* | *67.8* | *0.584* | *31* | *132* | *0.101* |
| *iPhone 11* | *1* | *SONY* | *2* | *0.696* | *67.2* | *0.593* | *277* | *378* | *0.494* |
| *SONY* | *2* | *iPhone 11* | *1* | *0.723* | *79.8* | *0.746* | *66* | *167* | *0.13* |
| *iPhone 11* | *2* | *SONY* | *2* | *0.585* | *60.2* | *0.515* | *39* | *140* | *0.109* |
| *SONY* | *2* | *iPhone 11* | *2* | *0.764* | *69* | *0.564* | *45* | *146* | *0.118* |

This algorithm was trained with a learning rate of 0.01 in all the previous sections. In these experiments, the best results were obtained when the source task images were purchased using SONY cameras under Illumination condition 2 and the new task images were purchased using iPhone 11 cameras under Illumination condition 1. A 20.128% improvement in mAP and a 11.142% improvement in accuracy were found compared to the next best results (obtained for source task of images purchased with iphone11camera under Illumination condition 1 and the new task's images purchased with the same camera under Illumination condition 1).

**Fine-tuning- freeze backbone (10 layers)**

Table19 - Transfer learning. Initial learning rate of 0.01, fine-tuning

| ***Pretrain set*** | | ***Retrain set*** | | ***mAP- train*** | ***Accuracy (%)*** | ***mAP- test*** | ***Converge epoch*** | ***Last training epoch*** | ***Time (hours) of training*** |
|---|---|---|---|---|---|---|---|---|---|
| *Camera type* | *Illumination condition* | *Camera type* | *Illumination condition* | | | | | | |
| *iPhone 11* | *1* | *iPhone 11* | *2* | *0.826* | *73.3* | *0.657* | *264* | *365* | *0.526* |
| *iPhone 11* | *2* | *iPhone 11* | *1* | *0.745* | *80.4* | *0.755* | *242* | *343* | *0.338* |
| *iPhone 11* | *1* | *SONY* | *2* | *0.76* | *74.8* | *0.6887* | *143* | *244* | *0.325* |
| *SONY* | *2* | *iPhone 11* | *1* | *0.814* | *82.6* | *0.802* | *400* | *400* | *0.392* |
| *iPhone 11* | *2* | *SONY* | *2* | *0.686* | *67.7* | *0.603* | *400* | *400* | *0.391* |
| *SONY* | *2* | *iPhone 11* | *2* | *0.826* | *72.9* | *0.65* | *199* | *300* | *0.299* |

The best results were obtained with the same camera and illumination conditions as Feature extraction. Since Fine-tuning involves retraining the fully connected layer and adjusting some of the convolution layers as well, it took 0.262 hours more than Feature extraction. However, the results are 7.506% better in mAP and 3.508% better in accuracy.

- Initial learning rate decreased to 0.0001:

**Feature extraction- freeze 24 convolution layers**

Table20 - Transfer learning. Initial learning rate of 0.0001, feature extraction

| ***Pretrain set*** | | ***Retrain set*** | | ***mAP-train*** | ***Accuracy (%)*** | ***mAP- test*** | ***Converge epoch*** | ***Last training epoch*** | ***Time (hours) of training*** |
|---|---|---|---|---|---|---|---|---|---|
| *Camera type* | *Illumination condition* | *Camera type* | *Illumination condition* | | | | | | |
| *iPhone 11* | *1* | *iPhone 11* | *2* | *0.726* | *69.4* | *0.591* | *1* | *102* | *0.138* |
| *iPhone 11* | *2* | *iPhone 11* | *1* | *0.659* | *68.2* | *0.595* | *2* | *103* | *0.079* |
| *iPhone 11* | *1* | *SONY* | *2* | *0.597* | *60.2* | *0.477* | *400* | *400* | *0.521* |
| *SONY* | *2* | *iPhone 11* | *1* | *0.712* | *78.2* | *0.719* | *16* | *117* | *0.092* |
| *iPhone 11* | *2* | *SONY* | *2* | *0.498* | *57.3* | *0.411* | *400* | *400* | *0.310* |
| *SONY* | *2* | *iPhone 11* | *2* | *0.764* | *69.8* | *0.57* | *400* | *400* | *0.324* |

The best results were obtained also when the source task images were purchased using SONY cameras under Illumination condition 2 and the new task images were purchased using iPhone 11 cameras under Illumination condition 1. These results are 3.75% less good on mAP index, and 2.05% on accuracy index in compared to the results of Feature extraction with learning rate of 0.01.

**Fine-tuning- freeze backbone (10 layers)**

Table21 - Transfer learning. Initial learning rate of 0.0001, fine-tuning

| ***Pretrain set*** | | ***Retrain set*** | | ***mAP- train*** | ***Accuracy (%)*** | ***mAP- test*** | ***Converge epoch*** | ***Last training epoch*** | ***Time (hours) of training*** |
|---|---|---|---|---|---|---|---|---|---|
| *Camera type* | *Illumination condition* | *Camera type* | *Illumination condition* | | | | | | |
| *iPhone 11* | *1* | *iPhone 11* | *2* | *0.747* | *71.2* | *0.598* | *184* | *285* | *0.399* |
| *iPhone 11* | *2* | *iPhone 11* | *1* | *0.657* | *67.4* | *0.591* | *2* | *103* | *0.100* |
| *iPhone 11* | *1* | *SONY* | *2* | *0.623* | *60.6* | *0.487* | *400* | *400* | *0.547* |
| *SONY* | *2* | *iPhone 11* | *1* | *0.736* | *79.4* | *0.744* | *226* | *327* | *0.324* |
| *iPhone 11* | *2* | *SONY* | *2* | *0.544* | *58.4* | *0.437* | *400* | *400* | *0.401* |
| *SONY* | *2* | *iPhone 11* | *2* | *0.779* | *69.8* | *0.601* | *400* | *400* | *0.409* |

With the same camera and illumination conditions as Feature extraction, the best results were achieved. With the same learning rate, the train took 0.232 hours longer than the feature extraction. There is a 3.477% improvement in mAP and a 1.53% improvement in accuracy.

# 5 Counting larvae outside the industrial ponds at day one

This chapter presents the development of a machine vision system for detecting and counting larvae on the day of their hatching, using color images acquired from a DSLR Nikon D510 camera is described. Evaluation of the experiments conducted and results from this study are detailed.

## 5.1 System design

The system developed aims to count larvae automatically on the first day they hatch, which is the most crucial day for production purposes. The images were acquired with a DSLR Nikon D3500 camera with the addition of a Nikon AF-S VR Micro-NIKKOR 105mm f/2.8G IF-ED macro lens. The camera shutter speed is 1: 2000 and the aperture value is 8. A sample of the larvae was manually removed from the growing ponds with a pump and placed in a petri dish 90 millimeters in diameter under the camera.

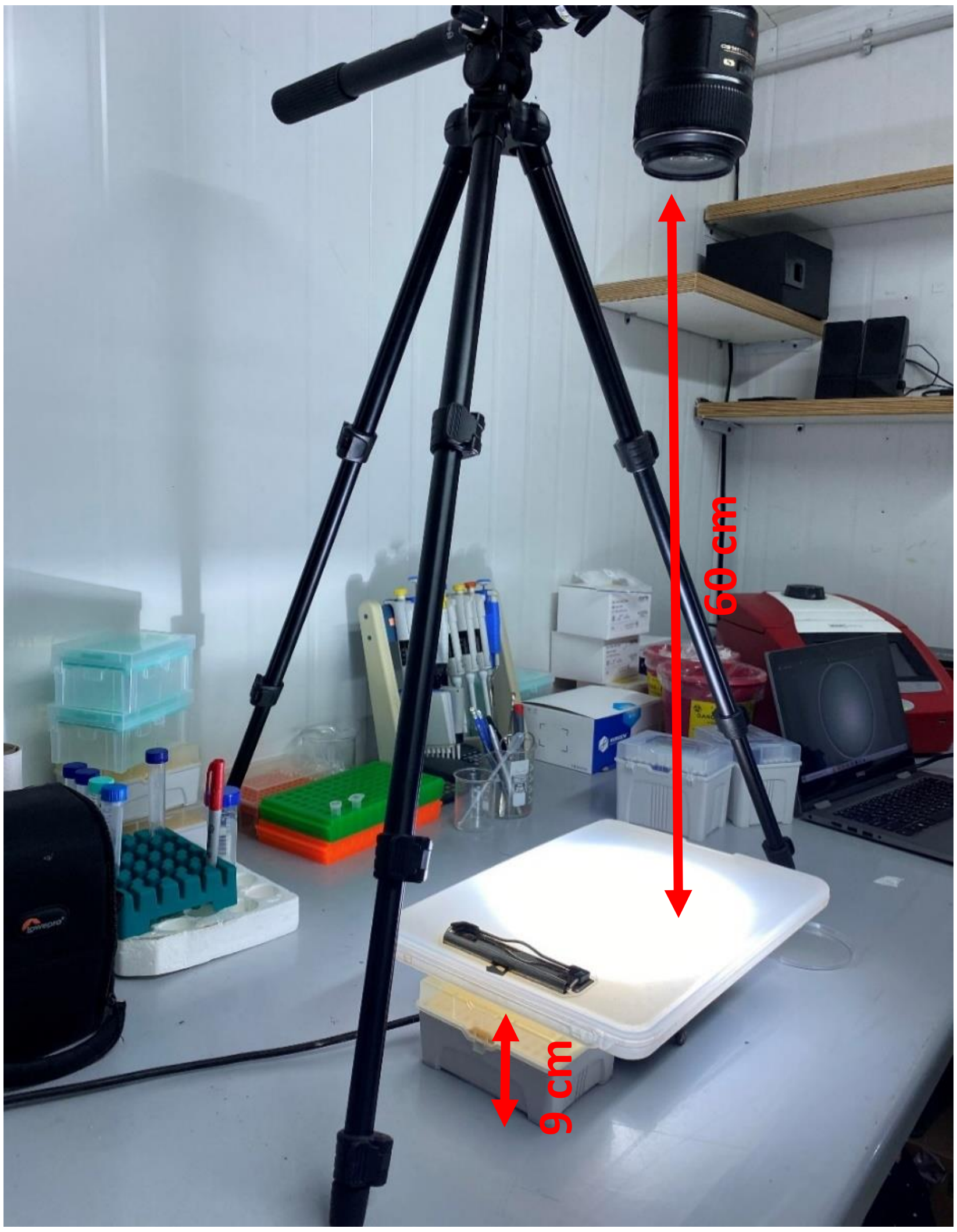

Figure 17- System design. The light high was 9 cm, and the camera distance was 60 cm from the light

## 5.2 Experimental design

The experiment was designed for larvae on the day of their hatching. The hatching of the larvae takes place in the initial industrial pond and immediately after their counting they are transferred to the industrial pond. All the larvae in the experiment were taken from the same initial industrial pond. seven different densities of larvae in the petri dish were tested (50, 100,150, 200, 300, 400, or 500 larvae). The petri dish was placed on a white background Above Lux of flashlight to maximize the quality of the images.

## 5.3 Data

200 images for each density level were acquired. Manual labelling was performed for 100 randomly selected images. The images resolution was 6000X4000. All images were acquired on day 1- at the first larvae's growth stage (Table 4), resulting in a larvae length of approximately 1.92 mm.

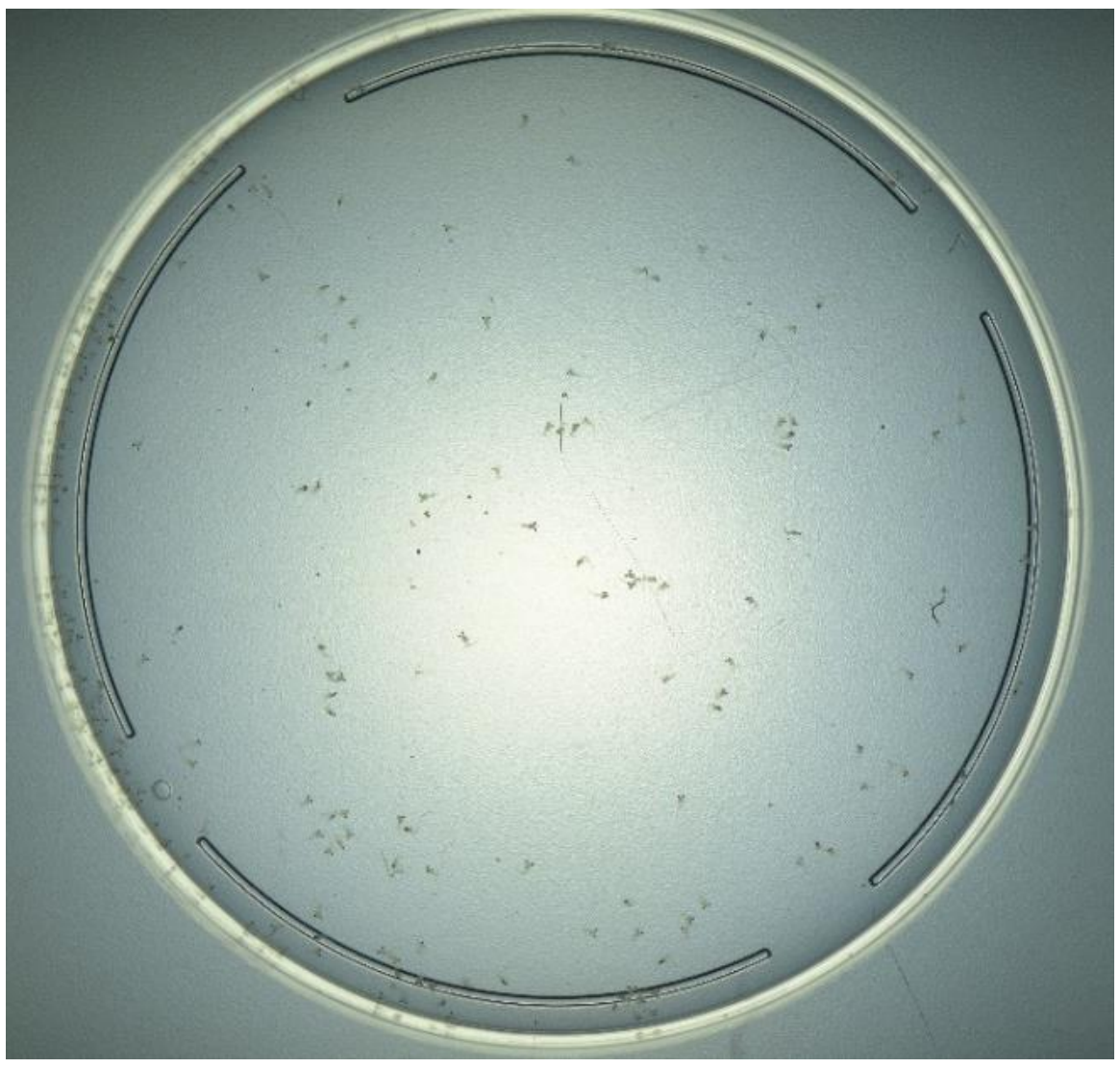

Figure 18- Example of an images from a density of 200

## 5.4 Results

The algorithm's results for each density were tested separately. Each density was trained individually with the images randomly divided into 80% images in the training set and 20% in the test set. Ten percent of the training set's images were used for validation. As a result, the object detection algorithm was trained and evaluated by using 72 training images, 8 validation images, and 20 test images. As expected, model performance falls off as larval density increases (Table 22).

Table 22- Average classification performance on each density for image size of 640, 300 epochs and 32 baches.

| Density | mAP-train set | mAP-test set | Accuracy (%) |
|---|---|---|---|
| **50** | 0.926 | 0.75 | 81.7 |
| **100** | 0.741 | 0.478 | 62.6 |
| **150** | 0.683 | 0.488 | 60.8 |
| **200** | 0.723 | 0.471 | 55.7 |
| **300** | 0.514 | 0.227 | 33.3 |
| **400** | 0.511 | 0.163 | 25.6 |
| **500** | 0.305 | 0.085 | 17.5 |

Considering that the training set of the original images may be too small, augmentation of $90^{\circ}$ rotations of the image was randomly performed on some of the images to increase the dataset and try to improve the model's performance. After augmentation the training set included 166 images, with a test set of 20 and validation set of 10 images.

Results (Table 23) revealed that the augmentations did not affect the results much nor consistently. In some densities the performance improved and in some were worse. The average and maximum difference in accuracy (%) between sequential densities was 8.183 and 16.4 respectively with maximum difference of 48.1 between the best case (50 density) and worst case (500 density).

Table 23- Average classification performance on each density after augmentation

| Density | mAP-train set | mAP-test set | Accuracy (%) |
|---|---|---|---|
| **50** | 0.845 | 0.534 | 66.2 |
| **100** | 0.761 | 0.5 | 63.5 |
| **150** | 0.8 | 0.525 | 64 |
| **200** | 0.745 | 0.425 | 53.7 |
| **300** | 0.538 | 0.24 | 37.3 |
| **400** | 0.59 | 0.182 | 26.9 |
| **500** | 0.35 | 0.099 | 18.1 |

## 5.5 Sensitivity analysis

**Transfer learning**

The model was trained on one density, then retrained on the other density using transfer learning techniques. Using the same hyperparameters (640-pixel image size, 300 epochs, 16 batches), 80 images were trained per density and 20 images were evaluated. Transfer learning was performed using fine tuning (freeze backbone) with Initial learning rate of 0.0001.

Table 24- Transfer learning. Initial learning rate of 0.0001, fine-tuning, pretrain density-50

| ***Retrain set*** | ***Pretrain set*** | ***mAP- train*** | ***Accuracy (%)*** | ***mAP- test*** | ***Converge epoch*** | ***Last training epoch*** | ***Time (hours) of training*** |
|---|---|---|---|---|---|---|---|
| **50** | **100** | 0.514 | 39.4 | 0.231 | 300 | 300 | 0.489 |
| | **150** | 0.525 | 43.8 | 0.296 | 300 | 300 | 0.569 |
| | **200** | 0.507 | 39.6 | 0.281 | 300 | 300 | 0.947 |
| | **300** | 0.365 | 15.7 | 0.083 | 300 | 300 | 1.414 |
| | **400** | 0.243 | 18.5 | 0.064 | 4 | 105 | 0.557 |
| | **500** | 0.153 | 4.3 | 0.012 | 80 | 181 | 1.071 |

Table 25- Transfer learning. Initial learning rate of 0.0001, fine-tuning, pretrain density-100

| ***Retrain set*** | ***Pretrain set*** | ***mAP- train*** | ***Accuracy (%)*** | ***mAP- test*** | ***Converge epoch*** | ***Last training epoch*** | ***Time (hours) of training*** |
|---|---|---|---|---|---|---|---|
| **100** | **50** | 0.76 | 58.7 | 0.417 | 300 | 300 | 0.289 |
| | **150** | 0.647 | 55.2 | 0.425 | 119 | 220 | 0.455 |
| | **200** | 0.641 | 50.4 | 0.395 | 300 | 300 | 0.179 |
| | **300** | 0.482 | 24.8 | 0.147 | 300 | 300 | 0.252 |
| | **400** | 0.484 | 17.9 | 0.079 | 181 | 282 | 1.482 |
| | **500** | 0.231 | 7.3 | 0.025 | 300 | 300 | 0.345 |

Table 26- Transfer learning. Initial learning rate of 0.0001, fine-tuning, pretrain density-150

| ***Retrain set*** | ***Pretrain set*** | ***mAP- train*** | ***Accuracy (%)*** | ***mAP- test*** | ***Converge epoch*** | ***Last training epoch*** | ***Time (hours) of training*** |
|---|---|---|---|---|---|---|---|
| **150** | **50** | 0.607 | 53.4 | 0.348 | 300 | 300 | 0.059 |
| | **100** | 0.593 | 46 | 0.295 | 300 | 300 | 0.087 |
| | **200** | 0.607 | 42.9 | 0.323 | 102 | 203 | 0.121 |
| | **300** | 0.423 | 19 | 0.110 | 154 | 255 | 0.212 |
| | **400** | 0.387 | 20.5 | 0.077 | 100 | 102 | 0.111 |
| | **500** | 0.218 | 5.2 | 0.019 | 300 | 300 | 0.339 |

Table 27- Transfer learning. Initial learning rate of 0.0001, fine-tuning, pretrain density-200

| *Retrain set* | *Pretrain set* | *mAP-train* | *Accuracy (%)* | *mAP- test* | *Converge epoch* | *Last training epoch* | *Time (hours) of training* |
|---|---|---|---|---|---|---|---|
| **200** | **50** | 0.553 | 37.9 | 0.206 | 300 | 300 | 0.059 |
| | **100** | 0.533 | 38.7 | 0.236 | 300 | 300 | 0.087 |
| | **150** | 0.571 | 45.4 | 0.30 | 300 | 300 | 0.107 |
| | **300** | 0.516 | 24.4 | 0.142 | 172 | 273 | 0.225 |
| | **400** | 0.468 | 21.8 | 0.09 | 0 | 101 | 0.108 |
| | **500** | 0.248 | 7.9 | 0.029 | 300 | 300 | 0.35 |

Table 28- Transfer learning. Initial learning rate of 0.0001, fine-tuning, pretrain density- 300

| *Retrain set* | *Pretrain set* | *mAP-train* | *Accuracy (%)* | *mAP- test* | *Converge epoch* | *Last training epoch* | *Time (hours) of training* |
|---|---|---|---|---|---|---|---|
| **300** | **50** | 0.584 | 44.4 | 0.288 | 151 | 252 | 0.052 |
| | **100** | 0.546 | 40.8 | 0.263 | 300 | 300 | 0.087 |
| | **150** | 0.579 | 47.2 | 0.336 | 300 | 300 | 0.106 |
| | **200** | 0.649 | 53.7 | 0.43 | 300 | 300 | 0.178 |
| | **400** | 0.459 | 19.2 | 0.08 | 0 | 101 | 0.109 |
| | **500** | 0.245 | 9 | 0.034 | 300 | 300 | 0.345 |

Table 29- Transfer learning. Initial learning rate of 0.0001, fine-tuning, pretrain density-400

| *Retrain set* | *Pretrain set* | *mAP-train* | *Accuracy (%)* | *mAP- test* | *Converge epoch* | *Last training epoch* | *Time (hours) of training* |
|---|---|---|---|---|---|---|---|
| **400** | **50** | 0.629 | 44.5 | 0.275 | 300 | 300 | 0.06 |
| | **100** | 0.573 | 41.5 | 0.262 | 300 | 300 | 0.087 |
| | **150** | 0.52 | 44.6 | 0.306 | 300 | 300 | 0.108 |
| | **200** | 0.62 | 50.2 | 0.381 | 300 | 300 | 0.181 |
| | **300** | 0.433 | 24.2 | 0.14 | 300 | 300 | 0.249 |
| | **500** | 0.281 | 9.4 | 0.040 | 300 | 300 | 0.348 |

Table 30- Transfer learning. Initial learning rate of 0.0001, fine-tuning, pretrain density-500

| *Retrain set* | *Pretrain set* | *mAP-train* | *Accuracy (%)* | *mAP- test* | *Converge epoch* | *Last training epoch* | *Time (hours) of training* |
|---|---|---|---|---|---|---|---|
| **500** | **50** | 0.51 | 43.2 | 0.261 | 300 | 300 | 0.064 |
|  | **100** | 0.364 | 27.5 | 0.141 | 0 | 101 | 0.030 |
|  | **150** | 0.419 | 25.3 | 0.149 | 0 | 101 | 0.038 |
|  | **200** | 0.588 | 46.1 | 0.346 | 1 | 102 | 0.061 |
|  | **300** | 0.427 | 26.4 | 0.156 | 193 | 294 | 0.243 |
|  | **400** | 0.443 | 17.3 | 0.064 | 0 | 101 | 0.108 |

The model performance was not improved by transfer learning, and in most runs, all 300 epochs were required to reach convergence. Training on density 100 as a source task yields the best results for all other densities. As compared to training the next best density-150, we achieve better results at the lowest density (50) and higher density (500), by 9.925% and 40.38%, respectively. As new tasks - for training and testing - densities of 150 and 200 yield the best results. Furthermore, the results on the lower densities decrease as the density increases in the training set in the source task, while the results on the higher densities increase.

**Density combination**

There may be a need for more data to obtain better accuracy results. As a result, training was conducted when all images from all densities appeared in the validation, test, and training sets. The training set included 560 images, with a test set of 140 images. Each density's accuracy percentage was then calculated separately.

Table 31- Average classification performance – train on all densities, test on each density separately. (image size- 640, epochs- 300, baches- 32)

| Density | mAP-test set | Accuracy (%) |
|---|---|---|
| **50** | 0.801 | 86.0 |
| **100** | 0.591 | 69.6 |
| **150** | 0.648 | 72.3 |
| **200** | 0.651 | 70.8 |
| **300** | 0.384 | 48.8 |
| **400** | 0.321 | 40.8 |
| **500** | 0.23 | 30.9 |

Across all densities, the model performance results have improved significantly. As compared to training on each density separately, the biggest change in accuracy was observed at density 300 with an increase of 46.55%, and the smallest change was observed at density 50 with an increase of 5.263%. Overall, the high densities seem to have been more affected by training across all images. However, the best results remained for the lowest density - 50.

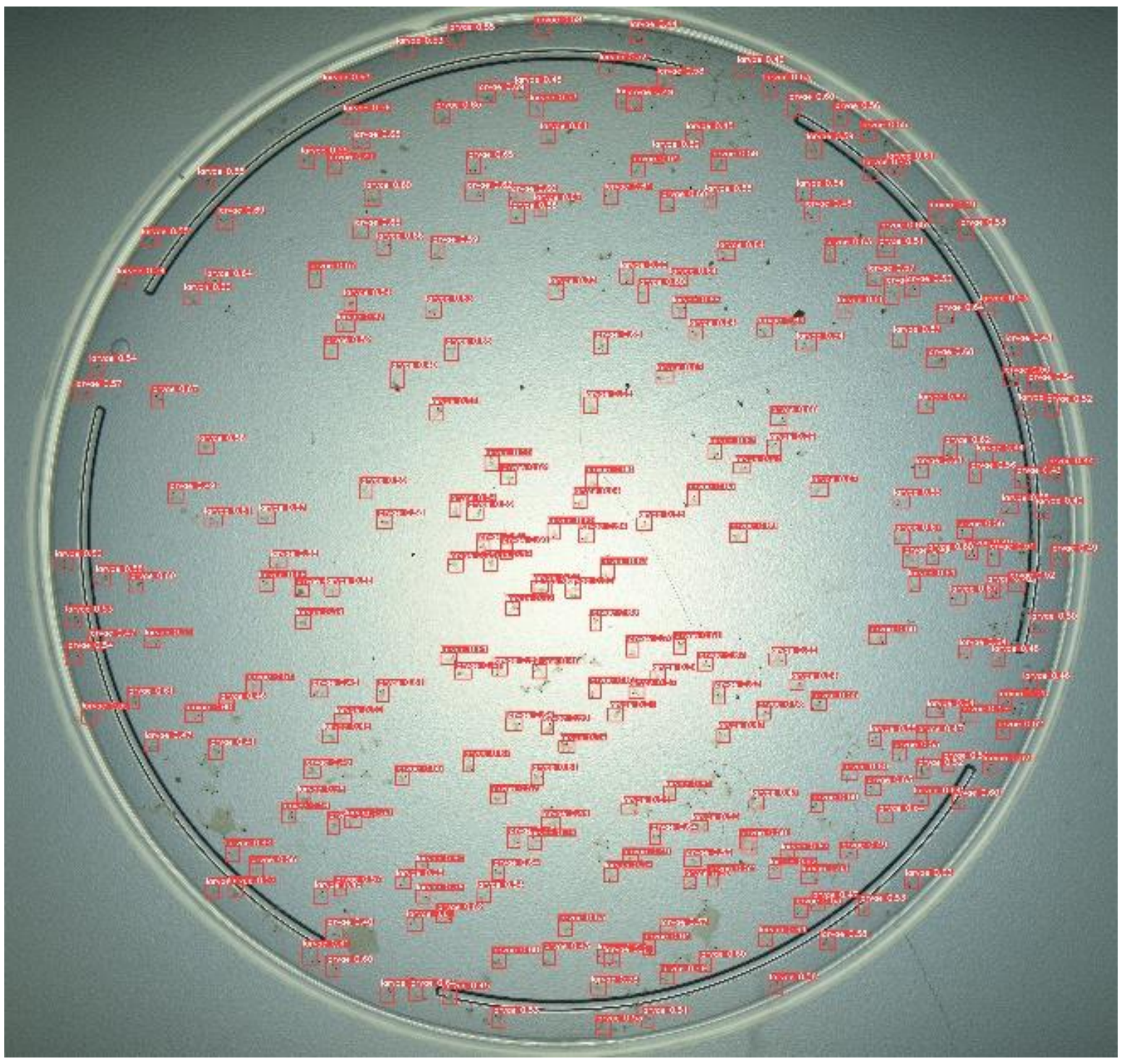

Figure 19- Test results on image of 300 density, with training on images of all densities

# 6 Growth function

## 6.1 Experimental design

The experiment in this study was conducted to measure the length of larvae according to their growth stage. All larvae that were used in the experiment were sourced from the same industrial pond located at Ben-Gurion University's life science lab. On a daily basis, several larvae were taken manually from the industrial pond and analyzed under a microscope -Nikon ECLIPSE Ci-S equipped with a Nikon DS-Fi3 RGB camera in resolution of 2880X2048 (10.8Appendix H - Nikon ECLIPSE Ci-S SPECS). The aim was to determine their growth stage. Once the growth stage was determined, an image or several images of the larva were acquired. The larva length was then manually measured based on the camera height and the resolution of each image.

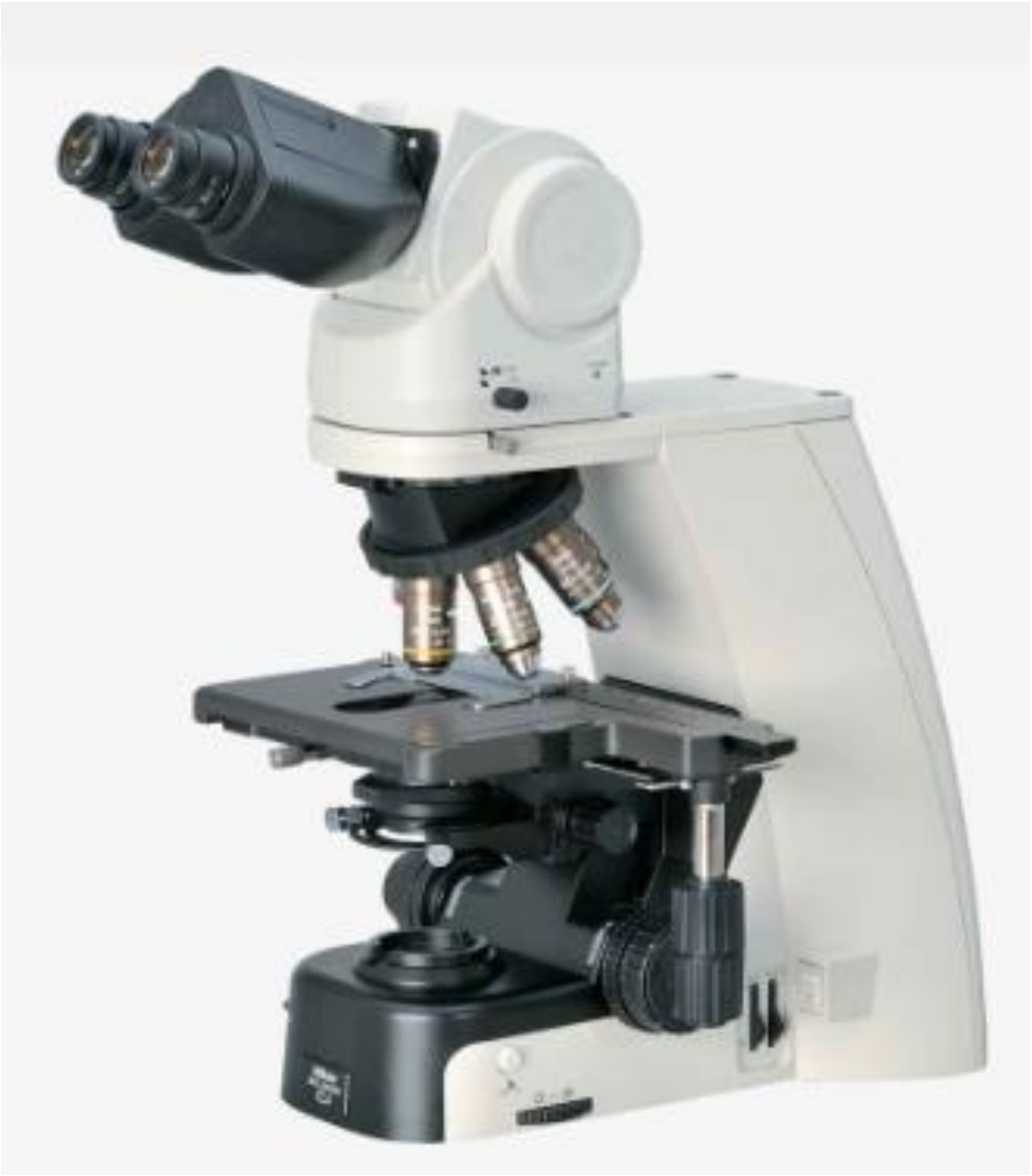

Figure 20- Nikon ECLIPSE Ci-S microscope

## 6.2 Data

A total of 10 larvae were sampled and photographed for each stage. A total of 290 images in resolution of 2880X2048 were acquired. 19 days were required for taking images from the first to the last stages.

## 6.3 Measurement

Each larva's length was measured from the image. Measurement was taken from eyes to the end of their tails. For each growth stage, the maximum, minimum, average, and standard deviation lengths were calculated. Below is a table (Table 32) showing the larvae length in millimeters for each stage in terms of average, standard deviation, maximum and minimum.

Table 32- Larvae growth stages

| **Growth-stage** | **Date** | **Age** | **Mean** | **Standard deviation** | **Maximum** | **Minimum** |
|---|---|---|---|---|---|---|
| **1** | 03/08 | 0 | 1.65 | 0.1 | 1.85 | 1.53 |
| **2** | 04/08 | 1 | 1.81 | 0.09 | 1.9 | 1.6 |
| **3** | 06/08-07/08 | 3 | 1.93 | 0.004 | 1.935 | 1.93 |
| **4** | 07/08 | 4 | 2.78 | 0.09 | 2.92 | 2.64 |
| **5** | 08/08 | 5 | 3.37 | 0.21 | 3.76 | 3.01 |
| **6** | 09/08 | 6 | 3.46 | 0.22 | 3.88 | 3.15 |
| **7** | 10/08-11/08 | 8 | 4.55 | 0.07 | 4.65 | 4.44 |
| **8** | 11/08-12/08 | 9 | 4.95 | 0.18 | 5.22 | 4.75 |
| **9** | 12/08-15/08 | 12 | 5.75 | 0.38 | 6.39 | 5.23 |
| **10** | 17/08 | 14 | 6.95 | 0.25 | 7.48 | 6.59 |
| **11** | 21/08 | 18 | 7.23 | 0.52 | 8.14 | 6.22 |

As larvae progress through the stages, their length increases with an average increase of 0.558 mm between two steps (std =0.41). It seems that in the first (1-3) and last (10-11) stages the increase is more moderate (Figure 21). Their body length ranges from 1.65-7.23 mm, meaning their length increases by 5.58 mm along the growth stages. Growth stage 11 have a smaller minimum length than stage 10 larvae. Since it has the largest standard deviation, with a gap of 0.14 from the next standard deviation (step 9), stage 11 has the largest variation in larval length between the maximum and minimum lengths, which is 1.92 mm. The jump between steps 9 and 10 is 1.2 inches, which is the largest jump between two consecutive steps. The smallest jump between two consecutive steps is 0.12 between steps 2 and 3.

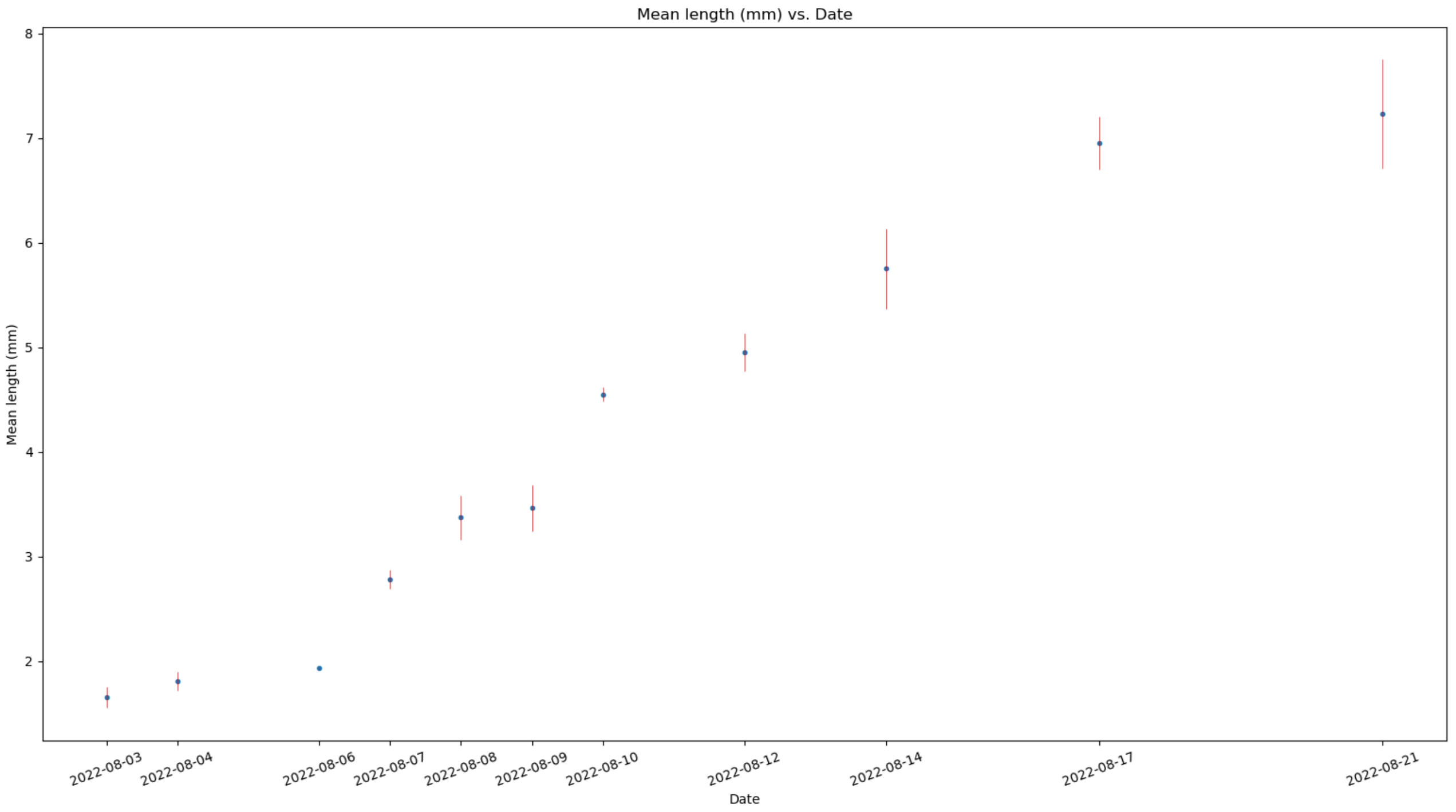

Figure 21- Mean length (mm) vs. Date

## 6.4 Growth model fitting results

A set of growth models was fitted to the same dataset, and the model that provided the most accurate fit was chosen. Model's selection was based on minimize the residual sum of squares (SSE). Goodness of fit is expressed by the coefficient of determination ($R^2$).

This study adopted five growth models, which are expressed as follows:

(1) VBGM- $L(t) = L_\infty(1 - \exp(-k_1(t - t_0)))$

(2) Gompertz- $L(t) = L_\infty e^{-k_2 e^{-a(t-t_r)}}$

(3) Linear - $L(t) = at + b$

(4) Power- $L(t) = at^b$

(5) Exponential - $L(t) = ae^{bt}$

Where, $t =$ age, $L(t) =$ Estimated mean length at age t and $L_\infty =$ Mean asymptotic length. In model number 1 $k_1$ is the Brody growth parameter, which is a relative growth rate parameter (with units yr-1) and $t_0$ is the hypothetical age when mean length is 0 (Zhu et al., 2009). In model 2 $k_2$ is the rate of exponential decrease of relative growth rate with age and $t_3$ is the initial stocking age. $a, b$, are constants (Tian et al., 1993). For each candidate model a curve was fitted so the data derived from our imaging at its parameters minimized the SSE.

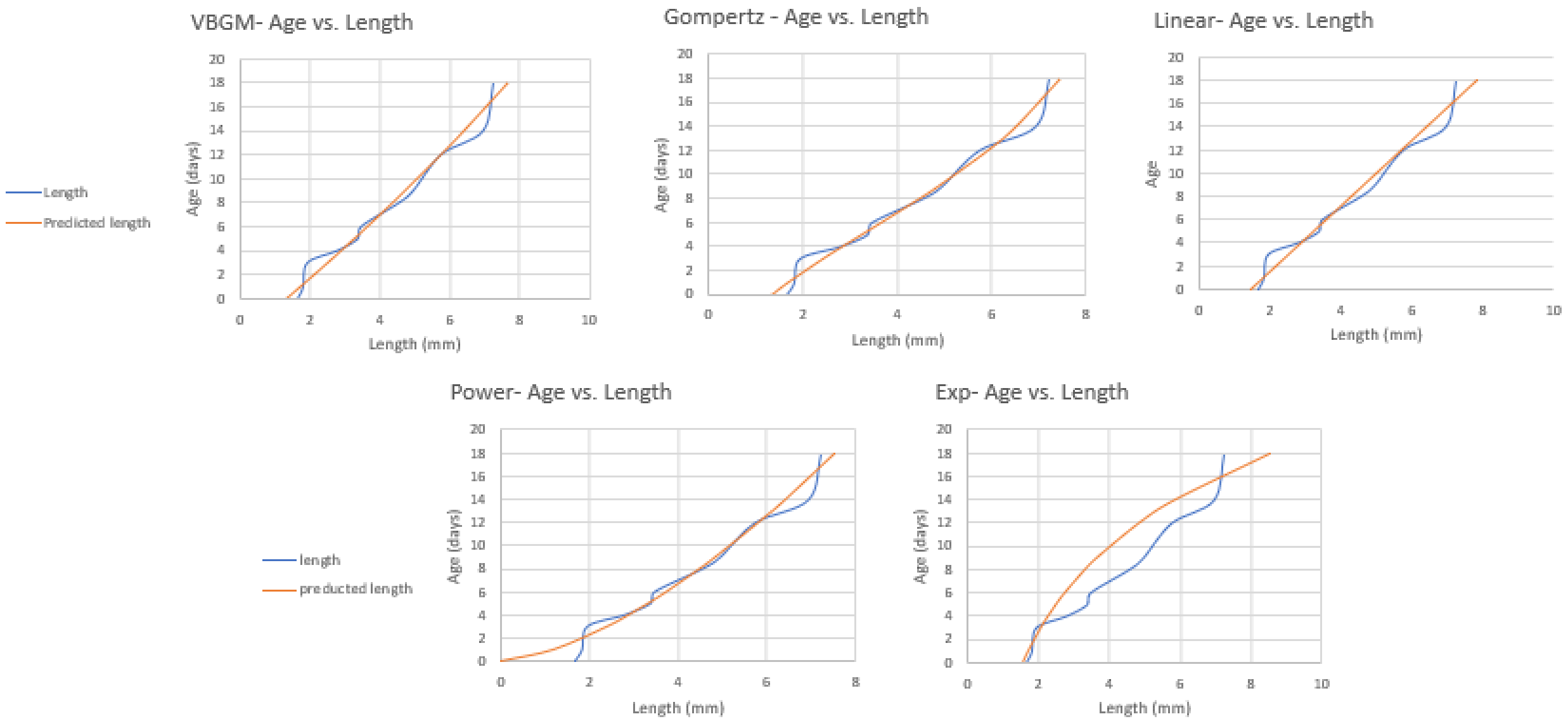

Figure 22- Growth curve for length-at-age data for larvae, fitted by VBGM, Gompertz, Linear, Power and Exponential from top left to bottom right.

After fitting the final equations obtained were as follows:

(1) VBGM- $L(t) = 27.217(1 - \exp(-0.0156(t + 3.229)))$

(2) Gompertz- $L(t) = 9.238e^{-1.933e^{-0.122(t+0.00004)}}$

(3) Linear - $L(t) = 0.352t - 1.479$

(4) Power- $L(t) = 1.191t^{0.638}$

(5) Exponential - $L(t) = 1.528e^{b0.095}$

The asymptotic length $L_\infty$ as estimated by VBGM was greater than that estimated by Gombertz by 17.979 mm.

According to Araya and Cubillos (Araya & Cubillos, 2006), the VBGM tended to estimate larger L values than the two-phase growth model based on length-at-age data sets of elasmobranch fish. In their view, growth should be analyzed with caution because VBGM does not always apply, and many species appear to function differently (Katsanevakis et al., 2008).

Goodness of fit was expressed by the coefficient of determination $R^2 = 1 - \frac{SSE}{SST}$ (Table 33). The closer $R^2$ gets to 1, the better the model.

Table 33- R^2 results for all five models

| ***Model*** | $R^2$ |
|---|---|
| VBGM | 0.973 |
| Gompertz | 0.983 |
| Linear | 0.969 |
| Power | 0.936 |
| Exponential | 0.867 |

Growth models fit to the length and age data show that the Gombertz model was selected as the best model with high correlation between days and length, followed by the VBGM (0.01% lower than Gombertz model), the linear model (0.004% lower than VBGM), power model (0.034% lower than liner model) and finally the exponential model (0.074% lower than power model). All four models except the exponential yielded $R^2$ greater than 0.9.

# 7 Discussion

## 7.1 Counting crustacean's larvae in industrial ponds at all larvae's growth stages

An entirely new system was developed in this study to count larvae under indoor growing conditions with controlled food, lighting, temperature, and salinity. The method is not dependent on a particular pond's features and can be applied to any pond. A user-friendly and inexpensive system that can potentially be used in different breeding facilities is provided by this system. Rather than using more advanced and professional cameras, a cell phone camera was deliberately used.

Under different illumination conditions and using different cameras, the system's performance was tested for 20 days in all growth stages of the larvae. Camera type and illumination conditions clearly affect the results. Using the iPhone11 camera under illumination condition number two (The lamps above the growth pond were turned on, but the room lighting was kept off) produced the best results.

Based on all the images from all the days, detection accuracy results of 88.4% and mAP of 0.855 were obtained. With this method, it is difficult to identify and count larvae in the first days of growth due to their small size and light color.

## 7.2 Counting crustacean's larvae outside the industrial ponds only at day one

A system developed in this chapter counts larvae only on the first day of hatching since the number of larvae on this day is most important for maintaining the breeding period. A professional camera is used to take images of larvae through a microscope after they have been manually counted. As larvae density increases, the algorithm's performance decreases.

As part of sensitivity analysis all images from all densities were trained and validated while each density's accuracy percentage was calculated separately. As with the previous chapter, this is a relatively simple system, although it is more expensive.

## 7.3 Growth function

This study developed a growth function for Macrobrachium Rosenberg larvae. In each of the 11 growth stages, 10 larvae were analyzed from specially acquired images acquired with a high-resolution camera mounted on a microscope. The image was used to measure the larva's length. A total of five models were fitted to the data and compared by $R^2$results. To describe Macrobrachium Rosenberg larvae growth, the equation must be biologically reasonable and statistically valid. For the experimental data, the functional forms of Gombertz and Von Bertalanffy provided the most accurate fit, with $R^2$ of 0.983 and 0.973, respectively. Gombertz's model is found to be more appropriate for this species than Von Bertalanffy, which is more commonly used in aquaculture.

# 8 Conclusion and future work

## 8.1 Counting crustacean's larvae in industrial ponds at all larvae's growth stages

The developed model inspired by "YOLOv5s" could count larvae in growth ponds. Images were obtained at Ben-Gurion University's life science lab during the 9-10 larval growth stage, when their size was between 6.07 mm and 7.05 mm. In this first experiment our model yielded a mAP metric of 0.961 (with an IoU of 0.5) and accuracy of 97%.

88.4% accuracy and 0.855 mAP were obtained in the second experiment which analyzed all larval growth stages using images acquired in the same lab for 20 days. The biggest challenge was the object's size. The smaller the larva was, the harder it was for the algorithm to identify it.

Future studies should focus on improving the accuracy of larval counts in the first stages of growth and evaluating whether the number of larvae in the bucket is accurate in estimating the total number of larvae in the pond.

## 8.2 Counting crustacean's larvae outside the industrial ponds only at day one

Develop a machine vision-based counting model for larvae on their first day of hatching. Among all densities, the lowest density achieved the best detection accuracy as expected (accuracy of of 81.7 and mAP of 0.75).

The training across all density methods resulted in an average recognition accuracy of 86%, which improved the accuracy of 50 density methods by 5.263%. Model implementation was performed using "YOLOv5s" algorithm with Stochastic Gradient Descent (SGD) loss function.

## 8.3 Growth function

For the first time, a growth function has been developed for Macrobrachium Rosenberg larvae. Growth is a continuous process influenced by internal and external factors, leading to individual and species-specific curves with different mathematical properties as they progress in their lives. The common practice among researchers who study fish growth is to a priori adopt the VBGM and base inference on this single model. However, in many cases, VBGM is not supported by the data and many species seem to follow different growth trajectories.

In this study, the performance of R2, was the best with the Gompertz model and achieved a score of 0.983.

# 9 References

Abe, S., Takagi, T., Takehara, K., Kimura, N., Hiraishi, T., Komeyama, K., Torisawa, S., & Asaumi, S. (2017). How many fish in a tank? Constructing an automated fish counting system by using PTV analysis. *Selected Papers from the 31st International Congress on High-Speed Imaging and Photonics*, *10328*(February 2017), 103281T. https://doi.org/10.1117/12.2270627

Antonucci, F., & Costa, C. (2020). Precision aquaculture: a short review on engineering innovations. *Aquaculture International*, *28*(1), 41–57. https://doi.org/10.1007/s10499-019-00443-w

Araya, M., & Cubillos, L. A. (2006). Evidence of two-phase growth in elasmobranchs. *Environmental Biology of Fishes*, *77*(3–4), 293–300. https://doi.org/10.1007/s10641-006-9110-8

Baer, A., Schulz, C., Traulsen, I., & Krieter, J. (2011). Analysing the growth of turbot (Psetta maxima) in a commercial recirculation system with the use of three different growth models. *Aquaculture International*, *19*(3), 497–511. https://doi.org/10.1007/s10499-010-9365-0

Brodzicki, A., Piekarski, M., Kucharski, D., Jaworek-Korjakowska, J., & Gorgon, M. (2020). Transfer Learning Methods as a New Approach in Computer Vision Tasks with Small Datasets. *Foundations of Computing and Decision Sciences*, *45*(3), 179–193. https://doi.org/10.2478/fcds-2020-0010

Cadieux, S., Michaud, F., & Lalonde, F. (2000). Intelligent system for automated fish sorting and counting. *IEEE International Conference on Intelligent Robots and Systems*, *2*(June), 1279–1284. https://doi.org/10.1109/IROS.2000.893195

Chang, S. K., Yuan, T. L., Hoyle, S. D., Farley, J. H., & Shiao, J. C. (2022). Growth Parameters and Spawning Season Estimation of Four Important Flyingfishes in the Kuroshio Current Off Taiwan and Implications From Comparisons With Global Studies. *Frontiers in Marine Science*, *8*(January). https://doi.org/10.3389/fmars.2021.747382

Chatain, B., Debas, L., & Bourdillon, A. (1996). A photographic larval fish counting technique: Comparison with other methods, statistical appraisal of the procedure and practical use. *Aquaculture*, *141*(1–2), 83–96. https://doi.org/10.1016/0044-8486(95)01206-0

Chiu, M. T., Xu, X., Wei, Y., Huang, Z., Schwing, A. G., Brunner, R., Khachatrian, H., Karapetyan, H., Dozier, I., Rose, G., Wilson, D., Tudor, A., Hovakimyan, N., Huang, T. S., & Shi, H. (2020). Agriculture-Vision: A Large Aerial Image Database for Agricultural Pattern Analysis. *2020 IEEE/CVF Conference on Computer Vision and Pattern Recognition (CVPR)*, 2825–2835. https://doi.org/10.1109/cvpr42600.2020.00290

Costa, C. S., Tetila, E. C., Astolfi, G., Sant'Ana, D. A., Brito Pache, M. C., Gonçalves, A. B., Garcia Zanoni, V. A., Picoli Nucci, H. H., Diemer, O., & Pistori, H. (2019). A computer vision system for oocyte counting using images captured by smartphone. *Aquacultural Engineering*, *87*(September). https://doi.org/10.1016/j.aquaeng.2019.102017

Deorankar, A. V., & Rohankar, A. A. (2020). An analytical approach for Soil and land classification system using image processing. *Proceedings of the 5th International Conference on Communication and Electronics Systems, ICCES 2020*, *Icces*, 1416–1420. https://doi.org/10.1109/ICCES48766.2020.09137952

Errami, A., & Khaldoun, M. (2018). Real Time Video Processing using RGB Remote Sensing by Drone. *2018 International Conference on Electronics, Control, Optimization and Computer Science (ICECOCS)*.

FAO. (2020). The State of World Fisheries and Aquaculture 2020. Sustainability in action. In *Fao*. https://doi.org/https://doi.org/10.4060/ca9229en

Farjon, G., Krikeb, O., Hillel, A. B., & Alchanatis, V. (2020). Detection and counting of flowers on apple trees for better chemical thinning decisions. *Precision Agriculture*, *21*(3), 503–521. https://doi.org/10.1007/s11119-019-09679-1

França Albuquerque, P. L., Garcia, V., da Silva Oliveira, A., Lewandowski, T., Detweiler, C., Gonçalves, A. B., Costa, C. S., Naka, M. H., & Pistori, H. (2019). Automatic live fingerlings counting using computer vision. *Computers and Electronics in Agriculture*, *167*(September), 105015. https://doi.org/10.1016/j.compag.2019.105015

Fuentes-Penailillo, F., Ortega-Farias, S., Fuente-Saiz, D. D. La, & Rivera, M. (2019). Digital count of Sunflower plants at emergence from very low altitude using UAV images. *IEEE CHILEAN Conference on Electrical, Electronics Engineering, Information and Communication Technologies, CHILECON 2019*, 1–5. https://doi.org/10.1109/CHILECON47746.2019.8988024

Geffen, O., Yitzhaky, Y., Barchilon, N., Druyan, S., & Halachmi, I. (2020). A machine vision system to detect and count laying hens in battery cages. *Animal*, *14*(12), 2628–2634. https://doi.org/10.1017/S1751731120001676

Gobalakrishnan, N., Pradeep, K., Raman, C. J., Ali, L. J., & Gopinath, M. P. (2020). A Systematic Review on Image Processing and Machine Learning Techniques for Detecting Plant Diseases. *Proceedings of the 2020 IEEE International Conference on Communication and Signal Processing, ICCSP 2020*, 465–468. https://doi.org/10.1109/ICCSP48568.2020.9182046

Gompertz, B. (1825). (1825). *On the nature of the function expressive of the law of human mortality and on a new mode of determining the value of life contingencies. Philosophical Transactions of the Royal Society of London*. 252–253.

Guo, Y., Shi, H., Kumar, A., Grauman, K., Rosing, T., & Feris, R. (2019). Spottune: Transfer learning through adaptive fine-tuning. *Proceedings of the IEEE Computer Society Conference on Computer Vision and Pattern Recognition*, *2019-June*, 4800–4809. https://doi.org/10.1109/CVPR.2019.00494

Halstead, M., McCool, C., Denman, S., Perez, T., & Fookes, C. (2018). Fruit Quantity and Ripeness Estimation Using a Robotic Vision System. *IEEE Robotics and Automation Letters*, *3*(4), 2995–3002. https://doi.org/10.1109/LRA.2018.2849514

Häni, N., Roy, P., & Isler, V. (2018). Apple Counting using Convolutional Neural Networks. *IEEE International Conference on Intelligent Robots and Systems*, *3*, 2559–2565. https://doi.org/10.1109/IROS.2018.8594304

Ibrahin, A. Kolo, J. G., Aibinu, B. M., James, A., Abdullahi, M. O. Folorunso, T. A. &Mutitu, A. A. (2017). A proposed fish counting algorithm using digital image processing technique. *ATBU Journal of Science, Technology and Education*, *5*(1), 9–15.

Kalantar, A., Edan, Y., Gur, A., & Klapp, I. (2020). A deep learning system for single and overall weight estimation of melons using unmanned aerial vehicle images. *Computers and Electronics in Agriculture*, *178*(August), 105748. https://doi.org/10.1016/j.compag.2020.105748

Kapach, K., Barnea, E., Mairon, R., Edan, Y., & Ben-Shahar, O. (2012). Computer vision for fruit harvesting robots - State of the art and challenges ahead. *International Journal of Computational Vision and Robotics*, *3*(1–2), 4–34. https://doi.org/10.1504/IJCVR.2012.046419

Katsanevakis, S., & Maravelias, C. D. (2008). Modelling fish growth: Multi-model inference as a better alternative to a priori using von Bertalanffy equation. *Fish and Fisheries*, *9*(2), 178–187. https://doi.org/10.1111/j.1467-2979.2008.00279.x

Khan, A., Khan, U., Waleed, M., Khan, A., Kamal, T., Marwat, S. N. K., Maqsood, M., & Aadil, F. (2018). Remote Sensing: An Automated Methodology for Olive Tree Detection and Counting in Satellite Images. *IEEE Access*, *6*, 77816–77828. https://doi.org/10.1109/ACCESS.2018.2884199

Kim, H., Lim, R., Seo, Y. Il, & Sheen, D. (2017). A modified von Bertalanffy growth model dependent on temperature and body size. *Mathematical Biosciences*, *294*(March 2016), 57–61. https://doi.org/10.1016/j.mbs.2017.10.006

Kirkwood, T. B. L. (2015). Deciphering death: A commentary on Gompertz (1825) 'On the nature of the function expressive of the law of human mortality, and on a new mode of determining the value of life contingencies.' *Philosophical Transactions of the Royal Society B: Biological Sciences*, *370*(1666). https://doi.org/10.1098/rstb.2014.0379

Kitano, B. T., Mendes, C. C. T., Geus, A. R., Oliveira, H. C., & Souza, J. R. (2019). Corn Plant Counting Using Deep Learning and UAV Images. *IEEE Geoscience and Remote Sensing Letters*, 1–5. https://doi.org/10.1109/lgrs.2019.2930549

Laradji, I., Rodriguez, P., Kalaitzis, F., Vazquez, D., Young, R., Davey, E., & Lacoste, A. (2020). *Counting Cows: Tracking Illegal Cattle Ranching From High-Resolution Satellite Imagery*. 1–6. http://arxiv.org/abs/2011.07369

Le, J., & Xu, L. (2017). An Automated Fish Counting Algorithm in Aquaculture Based on Image Processing. *2016 International Forum on Mechanical, Control and Automation (IFMCA 2016)*, *113*(Ifmca 2016), 358–366. https://doi.org/10.2991/ifmca-16.2017.56

Li, Wang, Z., Wu, S., Miao, Z., Du, L., & Duan, Y. (2020). Automatic recognition methods of fish feeding behavior in aquaculture: A review. *Aquaculture*, *528*(May), 735508. https://doi.org/10.1016/j.aquaculture.2020.735508

Li, Y., Jia, J., Zhang, L., Khattak, A. M., Sun, S., Gao, W., & Wang, M. (2019). Soybean seed counting based on pod image using two-column convolution neural network. *IEEE Access*, *7*, 64177–64185. https://doi.org/10.1109/ACCESS.2019.2916931

Lu, Y., & Young, S. (2020). A survey of public datasets for computer vision tasks in precision agriculture. *Computers and Electronics in Agriculture*, *178*(July), 105760. https://doi.org/10.1016/j.compag.2020.105760

Lugert, V., Thaller, G., Tetens, J., Schulz, C., & Krieter, J. (2016). A review on fish growth calculation: Multiple functions in fish production and their specific application. *Reviews in Aquaculture*, *8*(1), 30–42. https://doi.org/10.1111/raq.12071

Luo, S., Li, X., Wang, D., Li, J., & Sun, C. (2016). Automatic Fish Recognition and Counting in Video Footage of Fishery Operations. *Proceedings - 2015 International Conference on Computational Intelligence and Communication Networks, CICN 2015*, 296–299. https://doi.org/10.1109/CICN.2015.66

Manohar, & Gowda. (2020). Image Processing System based Identification and Classification of Leaf Disease: A Case Study on Paddy Leaf. *Proceedings of the International Conference on Electronics and Sustainable Communication Systems, ICESC 2020*, *Icesc*, 451–457. https://doi.org/10.1109/ICESC48915.2020.9155607

Mim, T. T., Sheikh, M. H., Shampa, R. A., Reza, M. S., & Islam, M. S. (2020). Leaves Diseases Detection of Tomato Using Image Processing. *2019 8th International Conference System Modeling and Advancement in Research Trends (SMART)*, 244–249. https://doi.org/10.1109/smart46866.2019.9117437

Mohammed Abd-Alsalam Selami, A., & Freidoon Fadhil, A. (2016). A Study of the Effects of Gaussian Noise on Image Features. *Kirkuk University Journal-Scientific Studies*, *11*(3), 152–169. https://doi.org/10.32894/kujss.2016.124648

Padilla, M. V. C., Arago, N. M., Damgo, J. C. M., Junio, G. D., Mamaril, J. J. R., Matienzo, R. N., & Natabio, G. S. (2019). Rice spikelet yield determination using image processing. *2018 IEEE 10th International Conference on Humanoid, Nanotechnology, Information Technology, Communication and Control, Environment and Management, HNICEM 2018*, 4–9. https://doi.org/10.1109/HNICEM.2018.8666312

Pedraza, I. L. A., Diaz, J. F. A., Pinto, R. M., Becker, M., & Tronco, M. L. (2019). Sweet Citrus Fruit Detection in Thermal Images Using Fuzzy Image Processing. *Communications in Computer and Information Science*, *1096 CCIS*, 182–193. https://doi.org/10.1007/978-3-030-36211-9_15

Plastiras, G., Kyrkou, C., & Theocharides, T. (2018). Efficient convnet-based object detection for unmanned aerial vehicles by selective tile processing. *ACM International Conference Proceeding Series*. https://doi.org/10.1145/3243394.3243692

Reddy, K. A., Reddy, N. V. M. C., & Sujatha, S. (2020). Precision Method for Pest Detection in Plants using the Clustering Algorithm in Image Processing. *Proceedings of the 2020 IEEE International Conference on Communication and Signal Processing, ICCSP 2020*, 894–897. https://doi.org/10.1109/ICCSP48568.2020.9182190

Redolfi, J. A., Felissia, S. F., Bernardi, E., Araguas, R. G., & Flesia, A. G. (2020). Learning to detect vegetation using computer vision and low-cost cameras. *Proceedings of the IEEE International Conference on Industrial Technology*, *2020-Febru*, 791–796. https://doi.org/10.1109/ICIT45562.2020.9067316

Ri, L. V., & Xvlqj, U. R. S. (2019). Yield Estimation of Chilli Crop using Image Processing Techniques. *IEEE Conference Proceedings (IEEE Conf Proc)*, *2020*(ICACCS), 200–204.

Tan, M., Pang, R., & Le, Q. V. (2019). *EfficientDet: Scalable and Efficient Object Detection*. http://arxiv.org/abs/1911.09070

Tian, H., Wang, T., Liu, Y., Qiao, X., & Li, Y. (2020). Computer vision technology in agricultural automation — A review. *Information Processing in Agriculture*, *7*(1), 1–19. https://doi.org/10.1016/j.inpa.2019.09.006

Tian, M., Guo, H., Chen, H., Wang, Q., Long, C., & Ma, Y. (2019). Automated pig counting using deep learning. *Computers and Electronics in Agriculture*, *163*(May), 104840. https://doi.org/10.1016/j.compag.2019.05.049

Tian, X., Leung, P. S., & Hochman, E. (1993). Shrimp growth functions and their economic implications. *Aquacultural Engineering*, *12*(2), 81–96. https://doi.org/10.1016/0144-8609(93)90018-7

Toh, Y. H., Ng, T. M., & Liew, B. K. (2009). Automated fish counting using image processing. *Proceedings -*

*2009 International Conference on Computational Intelligence and Software Engineering, CiSE 2009*. https://doi.org/10.1109/CISE.2009.5365104
Tombe, R. (2020). Computer Vision for Smart Farming and Sustainable Agriculture. *2020 IST-Africa Conference (IST-Africa)*, 8–15.
Wang, Z., Walsh, K. B., & Verma, B. (2017). On-tree mango fruit size estimation using RGB-D images. *Sensors (Switzerland)*, *17*(12), 1–15. https://doi.org/10.3390/s17122738
Xu, B., Wang, W., Falzon, G., Kwan, P., Guo, L., Chen, G., Tait, A., & Schneider, D. (2020). Automated cattle counting using Mask R-CNN in quadcopter vision system. *Computers and Electronics in Agriculture*, *171*(February), 105300. https://doi.org/10.1016/j.compag.2020.105300
Xu, B., Wang, W., Falzon, G., Kwan, P., Guo, L., Sun, Z., & Li, C. (2020). Livestock classification and counting in quadcopter aerial images using Mask R-CNN. *International Journal of Remote Sensing*, *41*(21), 8121–8142. https://doi.org/10.1080/01431161.2020.1734245
Y.Uno, K. chi. (1969). *Uno-1969-Larval development of Macrobrachium r (1).pdf* (p. 24).
Zhang, P., Zhang, X., Li, J., & Huang, G. (2006). The effects of body weight, temperature, salinity, pH, light intensity and feeding condition on lethal DO levels of whiteleg shrimp, Litopenaeus vannamei (Boone, 1931). *Aquaculture*, *256*(1–4), 579–587. https://doi.org/10.1016/j.aquaculture.2006.02.020
Zhu, L., Li, L., & Liang, Z. (2009). Comparison of six statistical approaches in the selection of appropriate fish growth models. *Chinese Journal of Oceanology and Limnology*, *27*(3), 457–467. https://doi.org/10.1007/s00343-009-9236-6
Zion, B. (2012). The use of computer vision technologies in aquaculture - A review. *Computers and Electronics in Agriculture*, *88*, 125–132. https://doi.org/10.1016/j.compag.2012.07.010

# 10 Appendices

## 10.1 Appendix A

Codes written in python used to train the CNN model, evaluate the validation set and test set:

monitor_val_score.py    evaluate_larvae.py    train_larvae.py

## 10.2 Appendix B

Code written in python for cropping the images before training:

crop_images_for_training.py

## 10.3 Appendix C

Code written in python for rescale the annotations:

annotation_rescale.py

## 10.4 Appendix D

Code written in python for finding the best threshold:

best_treshold.py

## 10.5 Appendix E - iPhone 11 camera SPEcS

- Dual 12MP Wide and Ultra-Wide cameras
- Wide: ƒ/1.8 aperture
- Ultra-Wide: ƒ/2.4 aperture and 120° field of view
- 2x optical zoom out
- Digital zoom up to 5x
- Portrait mode with advanced bokeh and Depth Control
- Portrait Lighting with six effects (Natural, Studio, Contour, Stage, Stage Mono, High-Key Mono)
- Optical image stabilization (Wide)
- Six-element lens (Wide); five-element lens (Ultra-Wide)
- True Tone flash with Slow Sync
- Panorama (up to 63MP)
- Sapphire crystal lens cover
- 100% Focus Pixels (Wide)
- Night mode (Wide)
- Deep Fusion (Wide)
- Next-generation Smart HDR for photos
- Wide color capture for photos and Live Photos
- Advanced red-eye correction
- Auto image stabilization
- Burst mode
- Photo geotagging
- Image formats captured: HEIF and JPEG

## 10.6 Appendix F - SONY DSC-HX90V camera SPECS

- CONTINUOUS SHOOTING SPEED (MAXIMUM) (WITH MAX. RECORDING PIXELS)

10 fps (for up to 10 shots)

- SHOOTING MODE

Intelligent Auto; Superior Auto; Program Auto; Aperture Priority; Shutter Speed Priority; Manual Exposure; MR (Memory Recall) 1,2,3; Movie Mode (Program Auto, Aperture Priority, Shutter Speed Priority, Manual Exposure); Panorama; Scene Selection

- STEADYSHOT (STILL IMAGE)

Optical SteadyShot

- FOCUS MODE

Single-shot AF, Continuous AF, DMF, Manual Focus

- ISO SENSITIVITY (MOVIE)

ISO 80–3200

- FOCUS TYPE

Contrast detection AF

- SHUTTER SPEED

iAuto (4 in – 1/2000) / Program Auto (1 in – 1/2000) / Aperture Priority (8 in – 1/2000) / Shutter Priority (30 in – 1/2000) / Manual (30 in – 1/2000)

- SCENE SELECTION

Portrait; Advanced Sports Shooting; Landscape; Sunset; Night Scene; Handheld Twilight; Night Portrait; Anti Motion Blur; Pet Mode; Gourmet; Beach; Snow; Fireworks; Soft Skin; High Sensitivity

- LIGHT METERING MODE

Multi Pattern; Center Weighted; Spot

- MINIMUM ILLUMINATION

Movie: Auto 6 lux (shutter speed 1/30)

## 10.7 Appendix G - Hyper parameters tuning

| *image-size* | *epochs* | *batch* | *conf* | *mAP-train* | *Accuracy (%)* | *FP / Total number of instances (%)* | *TP* | *FP* | *mAP- test* |
|---|---|---|---|---|---|---|---|---|---|
| 416 | 50 | 16 | 0.4 | 0.307 | 39.7 | 29 | 763 | 558 | *0.259678841* |
| 416 | 100 | 16 | 0.4 | 0.515 | 52.9 | 21.1 | 1017 | 406 | *0.438043237* |
| 416 | 150 | 16 | 0.4 | 0.596 | 62.9 | 19.2 | 1210 | 369 | *0.552751422* |
| 416 | 200 | 16 | 0.4 | 0.655 | 68.2 | 18.3 | 1311 | 351 | *0.61229074* |
| 416 | 300 | 16 | 0.4 | 0.713 | 71.7 | 16.3 | 1379 | 313 | *0.654881358* |
| 416 | 400 | 16 | 0.4 | 0.758 | 73.3 | 14.9 | 1410 | 286 | *0.672682166* |
| 416 | 500 | 16 | 0.4 | 0.789 | 75.6 | 13.5 | 1453 | 260 | *0.700492501* |
| 416 | 50 | 32 | 0.4 | 0.304 | 43.3 | 33.1 | 833 | 637 | *0.279568791* |
| 416 | 100 | 32 | 0.4 | 0.547 | 59.3 | 25.4 | 1140 | 489 | *0.484320134* |
| 416 | 150 | 32 | 0.4 | 0.622 | 64.4 | 20.2 | 1238 | 389 | *0.555470228* |
| 416 | 200 | 32 | 0.4 | 0.678 | 69.2 | 18.9 | 1330 | 364 | *0.612830937* |
| 416 | 300 | 32 | 0.4 | 0.728 | 71.8 | 16.1 | 1381 | 309 | *0.65068543* |
| 416 | 400 | 32 | 0.4 | 0.769 | 74.1 | 14.4 | 1425 | 276 | *0.684421659* |
| 416 | 500 | 32 | 0.4 | 0.796 | 76.4 | 12.5 | 1470 | 241 | *0.71620506* |
| 416 | 50 | 64 | 0.4 | 0.193 | 42.5 | 58.9 | 818 | 1133 | *0.225116357* |
| 416 | 100 | 64 | 0.4 | 0.52 | 57.6 | 22.3 | 1108 | 429 | *0.479879081* |
| 416 | 150 | 64 | 0.4 | 0.614 | 63.7 | 20.1 | 1225 | 387 | *0.553708076* |
| 416 | 200 | 64 | 0.4 | 0.663 | 66.8 | 18.8 | 1285 | 361 | *0.589071393* |
| 416 | 300 | 64 | 0.4 | 0.719 | 71.9 | 15.8 | 1383 | 304 | *0.650960922* |
| 416 | 400 | 64 | 0.4 | 0.76 | 74.5 | 14.2 | 1432 | 273 | *0.688647687* |
| 416 | 500 | 64 | 0.4 | 0.793 | 75.9 | 13.6 | 1459 | 261 | *0.698826611* |
| 416 | 50 | 128 | 0.4 | 0.0647 | 72.4 | 52.1 | 1392 | 1002 | *0.577657223* |
| 416 | 100 | 128 | 0.4 | 0.332 | 48.4 | 43.1 | 930 | 828 | *0.332371294* |
| 416 | 150 | 128 | 0.4 | 0.534 | 57.2 | 21.3 | 1099 | 410 | *0.46732989* |
| 416 | 200 | 128 | 0.4 | 0.557 | 60.9 | 22.3 | 1172 | 429 | *0.500351608* |
| 416 | 300 | 128 | 0.4 | 0.63 | 66.3 | 21.1 | 1274 | 405 | *0.5756194* |
| 416 | 400 | 128 | 0.4 | 0.68 | 68.9 | 19.2 | 1325 | 369 | *0.609213829* |
| 416 | 500 | 128 | 0.4 | 0.711 | 72.4 | 16.5 | 1392 | 317 | *0.65592891* |
| 416 | 50 | 256 | 0.4 | 0.000368 | 0.6 | 207.4 | 11 | 3988 | *1.50685E-05* |
| 416 | 100 | 256 | 0.4 | 0.115 | 39.4 | 104.3 | 757 | 2006 | *0.142944843* |
| 416 | 150 | 256 | 0.4 | 0.308 | 55.5 | 58.2 | 1068 | 1120 | *0.369953483* |
| 416 | 200 | 256 | 0.4 | 0.516 | 60.6 | 28.1 | 1166 | 541 | *0.493120223* |
| 416 | 300 | 256 | 0.4 | 0.601 | 66.3 | 19.9 | 1275 | 383 | *0.57387948* |
| 416 | 400 | 256 | 0.4 | 0.702 | 70.5 | 16.5 | 1356 | 317 | *0.631911337* |
| 416 | 500 | 256 | 0.4 | 0.731 | 72.5 | 14.6 | 1394 | 280 | *0.664155006* |
| 640 | 50 | 16 | 0.4 | 0.652 | 71.9 | 23.8 | 1382 | 457 | *0.61919862* |
| 640 | 100 | 16 | 0.4 | 0.732 | 76.7 | 22.6 | 1474 | 435 | *0.669350863* |
| 640 | 150 | 16 | 0.4 | 0.786 | 82.7 | 16.3 | 1591 | 314 | *0.756062567* |
| 640 | 200 | 16 | 0.4 | 0.824 | 85.9 | 13.9 | 1651 | 267 | *0.809545159* |
| 640 | 300 | 16 | 0.4 | 0.903 | 87.4 | 10.5 | 1681 | 202 | *0.832148671* |
| 640 | 400 | 16 | 0.4 | 0.907 | 90.5 | 7.9 | 1740 | 152 | *0.875006139* |
| 640 | 500 | 16 | 0.4 | 0.926 | 91.2 | 6.8 | 1754 | 130 | *0.885743916* |
| 640 | 50 | 32 | 0.4 | 0.585 | 68.3 | 30 | 1314 | 576 | *0.570872009* |
| 640 | 100 | 32 | 0.4 | 0.763 | 80.1 | 16.8 | 1540 | 323 | *0.723928869* |
| 640 | 150 | 32 | 0.4 | 0.854 | 87 | 12.9 | 1673 | 248 | *0.822167397* |
| 640 | 200 | 32 | 0.4 | 0.903 | 88.5 | 9.5 | 1701 | 183 | *0.846036792* |
| 640 | 300 | 32 | 0.4 | 0.916 | 90 | 8.9 | 1730 | 171 | *0.864170313* |
| 640 | 400 | 32 | 0.4 | 0.931 | 91.2 | 7.7 | 1754 | 149 | *0.882190943* |
| 640 | 500 | 32 | 0.4 | 0.933 | 91.4 | 7.5 | 1757 | 145 | *0.885207713* |
| 640 | 50 | 64 | 0.4 | 0.5 | 73.4 | 48 | 1412 | 923 | *0.601422071* |
| 640 | 100 | 64 | 0.4 | 0.796 | 83.3 | 17.2 | 1601 | 330 | *0.764219046* |
| 640 | 150 | 64 | 0.4 | 0.871 | 86.8 | 12.5 | 1669 | 241 | *0.82770586* |
| 640 | 200 | 64 | 0.4 | 0.892 | 87.9 | 10 | 1690 | 193 | *0.843536735* |
| 640 | 300 | 64 | 0.4 | 0.923 | 89.9 | 8.6 | 1728 | 165 | *0.867028356* |
| 640 | 400 | 64 | 0.4 | 0.93 | 91.2 | 7.8 | 1753 | 150 | *0.881261826* |
| 640 | 500 | 64 | 0.4 | 0.933 | 91 | 7.6 | 1750 | 146 | *0.883463383* |
| 1024 | 50 | 16 | 0.4 | 0.608 | 74.3 | 22.7 | 1429 | 437 | *0.640314221* |
| 1024 | 100 | 16 | 0.4 | 0.839 | 86.2 | 12.5 | 1658 | 241 | *0.813447893* |
| 1024 | 150 | 16 | 0.4 | 0.933 | 94.1 | 5.6 | 1810 | 107 | *0.923685431* |
| 1024 | 200 | 16 | 0.4 | 0.946 | 93.8 | 6 | 1804 | 115 | *0.916908205* |
| 1024 | 300 | 16 | 0.4 | 0.948 | 94.8 | 4.6 | 1823 | 88 | *0.931105971* |
| 1024 | 400 | 16 | 0.4 | 0.955 | 95.7 | 3.8 | 1841 | 74 | *0.948480308* |
| 1024 | 500 | 16 | 0.4 | 0.945 | 94.3 | 4.5 | 1814 | 87 | *0.929601073* |
| 1024 | 50 | 32 | 0.4 | 0.85 | 92.6 | 10.8 | 1781 | 207 | *0.898473918* |
| 1024 | 100 | 32 | 0.4 | 0.93 | 92.1 | 9 | 1772 | 173 | *0.89481014* |
| 1024 | 150 | 32 | 0.4 | 0.884 | 89.4 | 8.7 | 1720 | 167 | *0.863170862* |
| 1024 | 200 | 32 | 0.4 | 0.952 | 95.3 | 4.6 | 1833 | 88 | *0.952* |
| 1024 | 300 | 32 | 0.4 | 0.957 | 96.3 | 4.4 | 1851 | 84 | *0.950910926* |
| 1024 | 400 | 32 | 0.4 | 0.951 | 94.8 | 3.4 | 1823 | 65 | *0.935987353* |
| 1024 | 500 | 32 | 0.4 | 0.958 | 95.8 | 4 | 1843 | 76 | *0.950185001* |
| 1280 | 50 | 16 | 0.4 | 0.773 | 79.6 | 19 | 1531 | 365 | *0.717743099* |
| 1280 | 100 | 16 | 0.4 | 0.951 | 94.6 | 6.1 | 1820 | 117 | *0.93757683* |
| 1280 | 150 | 16 | 0.4 | 0.953 | 94.5 | 5.1 | 1818 | 98 | *0.932823062* |
| 1280 | 200 | 16 | 0.4 | 0.955 | 95.1 | 5.4 | 1828 | 103 | *0.934349358* |
| 1280 | 300 | 16 | 0.4 | 0.943 | 94.5 | 5.9 | 1818 | 113 | *0.930459917* |
| 1280 | 400 | 16 | 0.4 | 0.954 | 96 | 3.5 | 1846 | 67 | *0.950627387* |
| 1280 | 500 | 16 | 0.4 | 0.952 | 95.7 | 4.7 | 1840 | 91 | *0.941536844* |
| 1504 | 50 | 16 | 0.4 | 0.69 | 71.1 | 28.8 | 1368 | 554 | *0.560107112* |
| 1504 | 100 | 16 | 0.4 | 0.928 | 93.1 | 7.1 | 1791 | 136 | *0.911454797* |
| 1504 | 150 | 16 | 0.4 | 0.934 | 92.9 | 7.2 | 1787 | 139 | *0.909916997* |
| 1504 | 200 | 16 | 0.4 | 0.957 | 94.6 | 5.5 | 1820 | 105 | *0.932722092* |
| 1504 | 300 | 16 | 0.4 | 0.96 | 96.8 | 4.2 | 1861 | 80 | *0.959804952* |
| 1504 | 400 | 16 | 0.4 | 0.963 | 97 | 3.5 | 1865 | 68 | *0.961122811* |
| 1504 | 500 | 16 | 0.4 | 0.955 | 97 | 3.4 | 1866 | 66 | *0.963612258* |
| 1536 | 50 | 16 | 0.4 | 0.818 | 85.2 | 15.1 | 1638 | 291 | *0.77324003* |
| 1536 | 100 | 16 | 0.4 | 0.924 | 94.9 | 6.9 | 1825 | 132 | *0.93516469* |
| 1536 | 150 | 16 | 0.4 | 0.93 | 94.5 | 6.8 | 1818 | 131 | *0.923987448* |
| 1536 | 200 | 16 | 0.4 | 0.934 | 93.7 | 6.3 | 1802 | 122 | *0.91579783* |
| 1536 | 300 | 16 | 0.4 | 0.939 | 93.9 | 5.5 | 1806 | 105 | *0.923969865* |
| 1536 | 400 | 16 | 0.4 | 0.953 | 93.9 | 5.5 | 1806 | 105 | *0.923969865* |
| 1536 | 500 | 16 | 0.4 | 0.97 | 96.6 | 3.8 | 1858 | 73 | *0.955975652* |

## 10.8 Appendix H - Nikon ECLIPSE Ci-S SPECS

| | |
|---|---|
| Optical system | CFI60 Infinity Optical System |
| Illumination | 6V30W Halogen Lamp<br>Built-in ND4, ND8, NCB11 filters |
| Controls | ND filter IN/OUT switches |
| Eyepieces<br>(F.O.V. mm) | Sleeve diameter Φ30mm<br>• CFI 10X (22)<br>• CFI 12.5X (16)<br>• CFI 15X (14.5)<br>CFI UW 10X (25) |
| Focusing | Coaxial Coarse/Fine focusing, Focusing stroke: 30 mm, Coarse: 9.33 mm/rotation, Fine: 0.1 mm/rotation, Fine movement scale 1µm, Coarse motion torque adjustable, Refocusing function |
| Tubes<br>F.O.V. 22 mm<br>(Eyepiece/Port) | C-TB Binocular Tube<br>C-TE2 Ergonomic Binocular Tube (Eyepiece: Port = 100:0, 50:50) via optional C-TEP2 DSC Port, C-TEP3 |
| | DSC Port C-0.55X<br>Inclination angle: 10-30 degree, Extension: up to 40 mm |
| Tubes<br>F.O.V. 25 mm<br>(Eyepiece/Port) | F-CTF Trinocular Tube F (Eyepiece: Port = 100:0, 0:100)<br>TC-TT Trinocular Tube T (Eyepiece: Port = 100:0, 20:80, 0:100) |
| Nosepieces | C-N6 ESD Sextuple Nosepiece ESD<br>C-N6A Sextuple Nosepiece with Analyzer Slot |
| Stages | Cross travel 78 (X) × 54 (Y) mm, with vernier calibrations, stage handle height and torque adjustable for all stages<br>• C-SR2S Right Handle Stage with 2S Holder<br>• C-CSR Right Handle Ceramic-coated Stage<br>• C-CSR1S Right Handle Ceramic-coated Stage with 1S Holder<br>• C-H2L Specimen Holder 2L(Option)<br>• C-H1L Specimen Holder 1L (Option) |
| Condensers (NA)<br>Motorized | - |
| Condensers<br>(NA)<br>Manual | Focusing stroke: 27 mm<br>• C-AB Abbe Condenser<br>• C-AR Achromat Condenser<br>• C-DO Darkfield Condenser Oil<br>• C-DD Darkfield Condenser Dry<br>• C-PH Phase Contrast Turret Condenser<br>• C-AA Achromat/ Aplanat<br>• C-SA Slide Achromat Condenser 2-100X<br>• C-SW Swing-out Achromat Condenser 1-100X<br>• C-SWA Swing-out Achromat Condenser 2-100X<br>• C-LAR LWD Achromat Condenser |
| Observation methods[*3] | Brightfield, Epi-fluorescence, Darkfield, Phase contrast, Simple polarizing, Sensitive color polarizing |
| Epi-fluorescence attachment | CI-FL-2 Epi-fluorescence Attachment (4 filter cubes mountable)<br>D-FL-2 U-EPI Epi-fluorescence Attachment (6 filter cubes mountable, Terminator mechanism) |
| Epi-fluorescence light source | D-LEDI Fluorescence LED Illumination system<br>C-HGFI/HGFIE HG Precentered Fiber Illuminator Intensilight (130W) |
| Power consumption | 38W (Brightfield configuration) |
| Weight (approx.) | 13.4 kg (Binocular standard set) |

Results for “eclipse Ci” | Nikon Instruments Inc.